\documentclass[10pt,twocolumn,letterpaper]{article}

\usepackage[pagenumbers]{cvpr} 

\usepackage{graphicx}
\usepackage{amsmath}
\usepackage{amssymb}
\usepackage{booktabs}

\usepackage{bbding}

\usepackage[pagebackref,breaklinks,colorlinks]{hyperref}
\usepackage[accsupp]{axessibility}  

\usepackage[capitalize]{cleveref}
\crefname{section}{Sec.}{Secs.}
\Crefname{section}{Section}{Sections}
\Crefname{table}{Table}{Tables}
\crefname{table}{Tab.}{Tabs.}

\def\cvprPaperID{7524} 
\def\confName{CVPR}
\def\confYear{2022}

\begin{document}

\title{Learning Adaptive Warping for Real-World Rolling Shutter Correction}

\author{Mingdeng~Cao\textsuperscript{1} \quad
        Zhihang~Zhong\textsuperscript{2} \quad
        Jiahao~Wang\textsuperscript{1} \quad
        Yinqiang~Zheng\textsuperscript{2} \href{mailto:yqzheng@ai.u-tokyo.ac.jp}{\Envelope} \quad
        Yujiu~Yang\textsuperscript{1} \href{mailto:yang.yujiu@sz.tsinghua.edu.cn}{\Envelope} \\
\textsuperscript{1}Tsinghua Shenzhen International Graduate School, Tsinghua University \\
\textsuperscript{2}The University of Tokyo
}
\maketitle

\begin{abstract}
This paper proposes the first real-world rolling shutter (RS) correction dataset, BS-RSC, and a corresponding model to correct the RS frames in a distorted video. Mobile devices in the consumer market with CMOS-based sensors for video capture often result in rolling shutter effects when relative movements occur during the video acquisition process, calling for RS effect removal techniques. However, current state-of-the-art RS correction methods often fail to remove RS effects in real scenarios since the motions are various and hard to model. To address this issue, we propose a real-world RS correction dataset BS-RSC. Real distorted videos with corresponding ground truth are recorded simultaneously via a well-designed beam-splitter-based acquisition system. BS-RSC contains various motions of both camera and objects in dynamic scenes. Further, an RS correction model with adaptive warping is proposed. Our model can warp the learned RS features into global shutter counterparts adaptively with predicted multiple displacement fields. These warped features are aggregated and then reconstructed into high-quality global shutter frames in a coarse-to-fine strategy. Experimental results demonstrate the effectiveness of the proposed method, and our dataset can improve the model's ability to remove the RS effects in the real world. The project is available at \url{https://github.com/ljzycmd/BSRSC}.

\end{abstract}

\section{Introduction}
\label{sec:intro}
Most consumer cameras adopt CMOS sensors for imaging due to their low power consumption, compact design, and fast imaging. At the same time, most CMOS sensors have rolling shutter~(RS) effects during imaging. Unlike a global shutter (GS) camera capturing all pixels simultaneously, an RS camera sequentially captures the image pixels row by row. Therefore, the RS distortions would occur in the recorded images and videos when relative movements arise between the camera and objects. The RS distortions significantly impair the visual quality. Moreover, the distorted images and videos deteriorate the performance of some downstream tasks, like 3D reconstruction, pose estimation, and depth prediction~\cite{albl2015r6p, dai2016rolling, lao2020rolling, fan2021rs}, leading to erroneous, undesirable, and distorted results.

There are usually two ways to mitigate the performance gap of existing computer vision algorithms working on the RS distorted and GS images. The first is to keep the original RS images unchanged and adapt the algorithms to the RS distorted images. Thus, many RS-aware algorithms are proposed in 3D vision field, \eg, RS structure-from-motion reconstruction~\cite{hedborg2012rolling, zhuang2017rolling}, RS stereo~\cite{saurer2013rolling}, RS camera calibration~\cite{oth2013rolling} and RS absolute camera pose~\cite{albl2015r6p, albl2016rolling, albl2019rolling, lao2018rolling}. An arguable better way is to correct the RS distorted images into GS images. In this way, we don't need to modify existing vision algorithms and can obtain visual-friendly images. Therefore, correcting the rolling shutter~(RSC) images is increasingly becoming significant in photography and has attracted considerable research attention recently~\cite{rengarajan2017unrolling, liu2020deep, albl2020two, fan2021sunet}.

\begin{figure}[!t]\footnotesize
    \centering
    \begin{tabular}{@{\hspace{1mm}}c@{\hspace{0.5mm}}c@{\hspace{1mm}}}
        \includegraphics[width=0.48\linewidth]{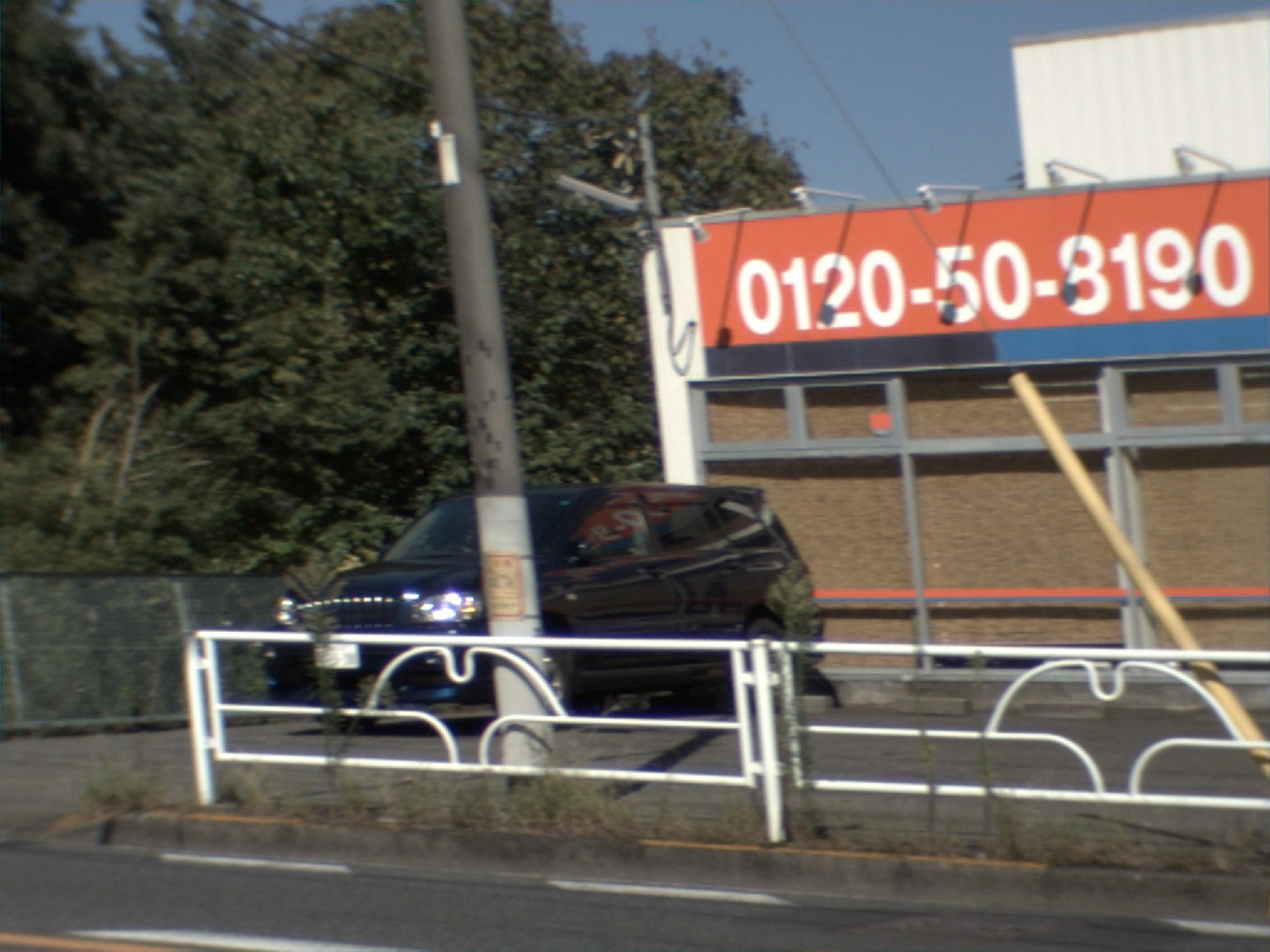} &
        \includegraphics[width=0.48\linewidth]{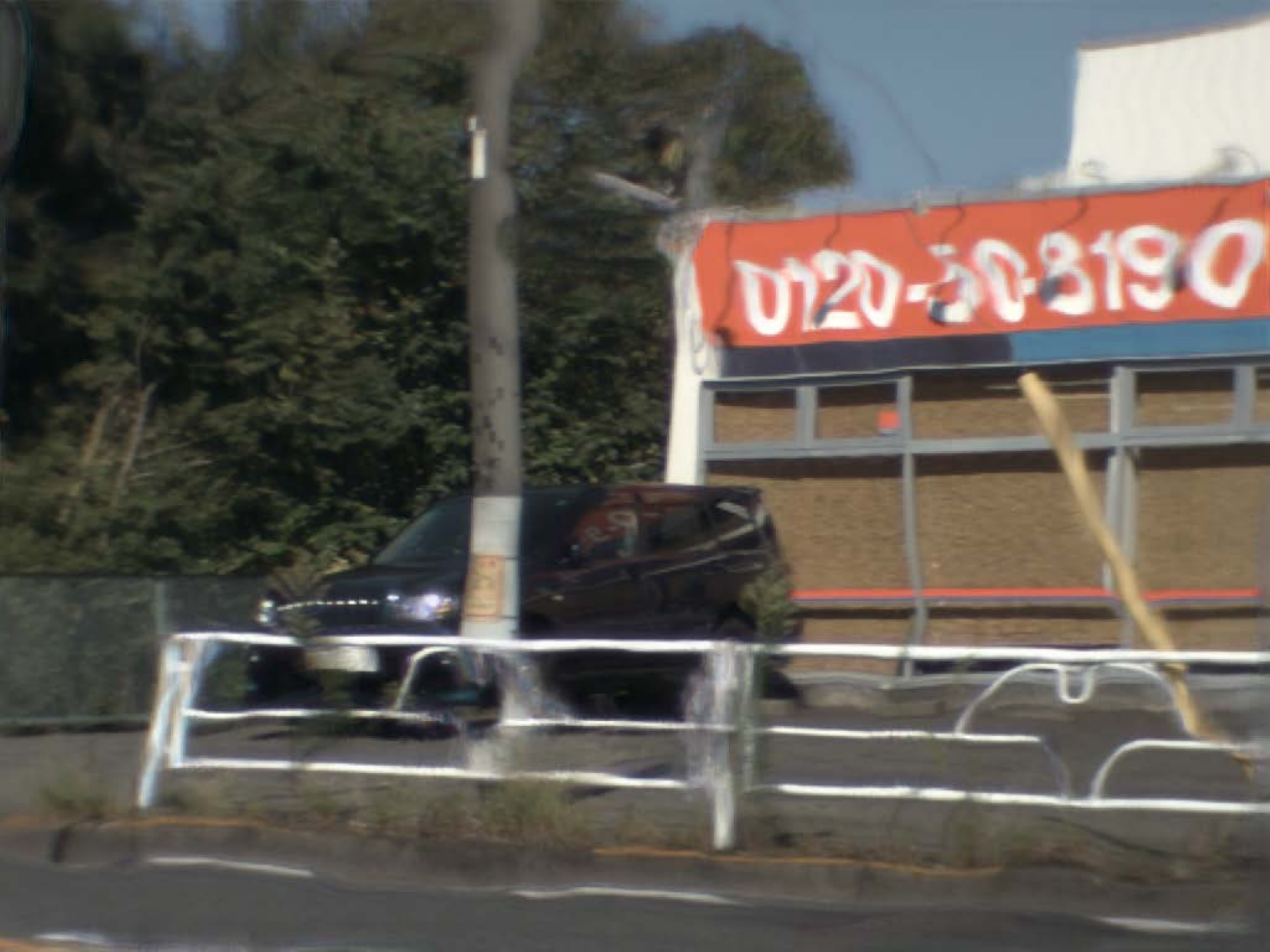}  \\
        (a) RS frame  & (b) DSUN~\cite{liu2020deep} \\
        \includegraphics[width=0.48\linewidth]{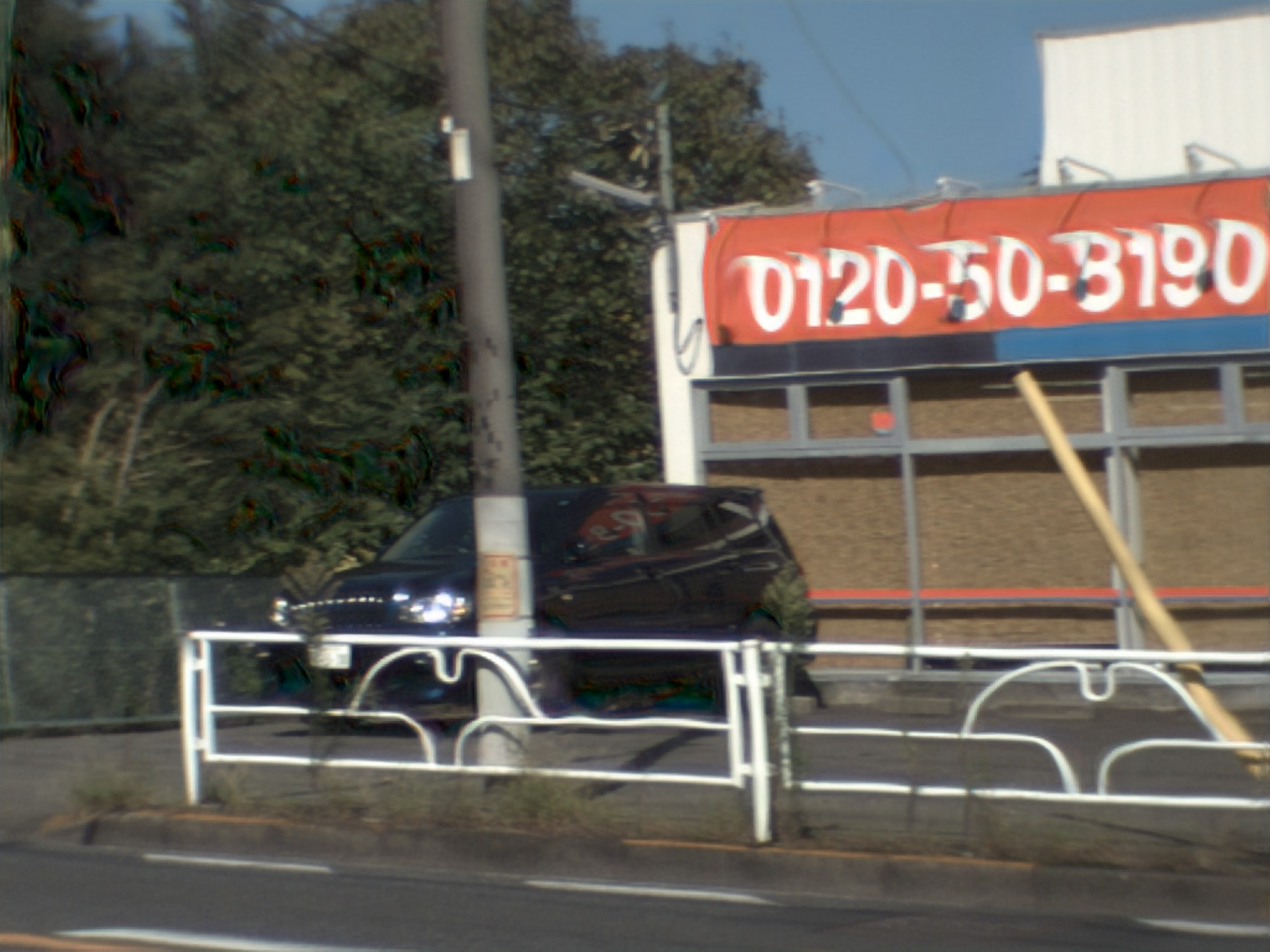} &
        \includegraphics[width=0.48\linewidth]{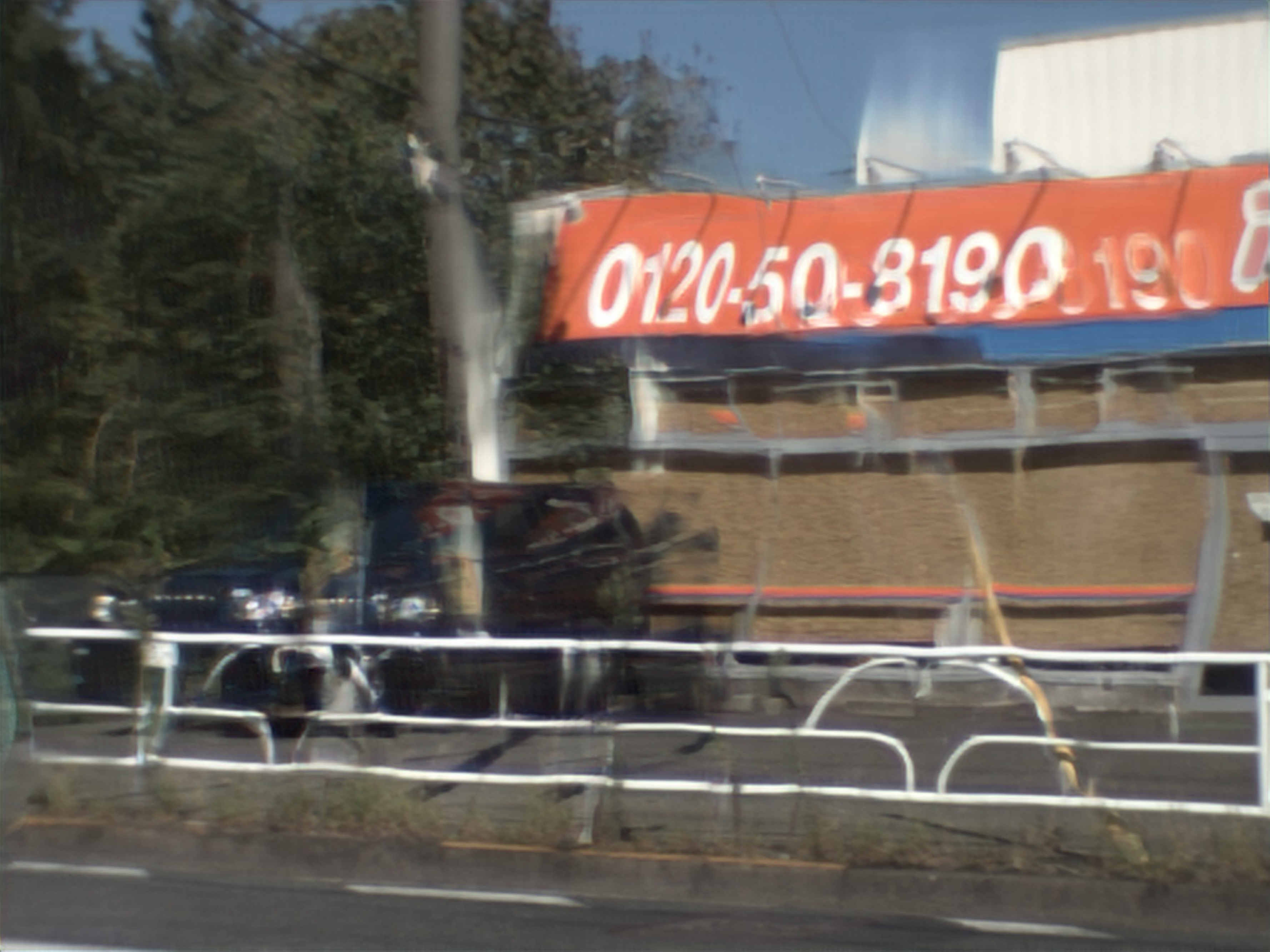}  \\
        (c) JCD~\cite{zhong2021towards}  &  (d) SUNet~\cite{fan2021sunet}\\
        \end{tabular}
        \caption{The real-world rolling shutter correction results of existing state-of-the-art methods trained with synthesized data. We see that all methods failed to remove the RS effects and even introduced many artifacts into the corrected frame.}%
    \label{fig:results_motivation}
    \vspace{-3mm}
\end{figure}

Existing RS effect removal methods can be categorized into single-image- and multi-frame-based. When restoring the GS image from only one RS image, many external constraints or priors (\eg, geometric priors) are adopted~\cite{rengarajan2016bows, rengarajan2017unrolling, lao2018robust, zhuang2019learning} since it is a highly ill-posed problem. Compared to single-image-based correction, multi-frame-based methods are more general and can utilize motion information for correction. Due to the great success of convolutional neural networks~(CNNs) on various computer vision tasks and the proposed synthesized RSC datasets, researchers designed specific model architectures to remove the RS distortions in an end-to-end manner based on multiple frames. Usually, the motions across multiple frames are modeled first. Then the GS image corresponding to the reference RS frame is restored by warping operations. For instance, Liu~\etal~\cite{liu2020deep} predict velocity field from the correlation volume, and Fan~\etal~\cite{fan2021sunet} utilize PWC-Net framework~\cite{sun2018pwc} to estimate the undistortion flow to correct the RS frame. They both adopt forward warping to remove the RS effect, and have achieved some promising results. However, the corrected GS images still suffer from blurs and texture detail loss for the following reasons: 1) The modeled motions are inaccurate since there is no ground truth for supervision during the training process. 2) The warping operations are not learnable, which cannot aggregate the features adaptively. 3) Meanwhile, some regions in the potential GS frame do not appear in the input RS frames. Thereby, it is difficult for the model to generate unseen areas. 4) Moreover, these models are trained on the synthesized RSC datasets where the motions are rather monotonous. And many artifacts exist in the synthesized RS frames, greatly restricting the model's performance on the natural RS image correction. \cref{fig:results_motivation} shows some real-world RSC results of state-of-the-art methods trained with synthesized data. 

To move beyond these limitations mentioned above, we propose a novel adaptive warping module and a real-world dataset for rolling shutter correction. Our model takes three consecutive frames as input, restoring the GS frame corresponding to the central RS frame at intermediate imaging time. We propose an adaptive warping module to better exploit high-quality GS frame restoration by mitigating inaccurate RS motion estimation and warping problems. Firstly, multi-scale features of each RS frame are extracted. Then, we construct a correlation volume to build the correspondence between central and neighboring RS features. The volume is used to predict multiple motion fields rather than only one generated in previous works~\cite{liu2020deep, fan2021sunet}. After that, an adaptive attention mechanism is proposed to warp the RS features by aggregating the contextual features according to the predicted motion fields. The designed warping process is learnable, aggregating the features to the GS-aware features attentively and adaptively. Note that we perform adaptive warping at all scales. A decoder network further decodes these warped multi-scale features and reconstructs the corresponding GS frame. The proposed model can be trained in an end-to-end manner. 

Considering the performance gap on the synthesized datasets and real RS distorted scenarios, we propose BS-RSC, the first dataset for real-world RSC with various motions in dynamic scenes, collected by a well-designed beam-splitter acquisition system. An RS camera and a GS camera are physically aligned to capture RS distorted and GS frames simultaneously. 

Our contributions can be summarized as follows:
\begin{itemize}
    \item We propose a novel feature warping module for rolling shutter correction, which adaptively warps RS features into global counterparts for high-quality GS frame restoration.
    \item We contribute BS-RSC, the first real-world RSC dataset~(devoid of motion blur) with various motions collected by a well-designed beam-splitter acquisition system, bridging the gap for real-world RSC task.
    \item The quantitative and qualitative experimental results on real-world and synthetic datasets show the excellent performance of the proposed method against the state-of-the-art methods.
\end{itemize}

\section{Related Works}
\subsection{Deep Rolling Shutter Correction}
CNNs are used to remove the RS effects due to the considerable success in many computer vision tasks. For single image RSC, Rengarajan~\etal~\cite{rengarajan2017unrolling} proposed a CNN architecture to estimate the row-wise camera motion from a single image and undo RS distortions back to the time of the first-row exposure. They adopted a long rectangular convolutional kernel to learn the effects produced by row-wise exposure specifically. Zhuang~\etal\cite{zhuang2019learning} further proposed a structure-and-motion-aware RS correction model that reasons about the concealed motions between the scanlines as well as the scene structure, where the camera scanline velocity and depth are estimated. 

Since single image RSC is a highly ill-posed task, multi-frame RSC can perform better by modeling the RS motion more accurately and has recently received much attention. Liu~\etal~\cite{liu2020deep} proposed an end-to-end network for RSC by predicting dense displacement field from two consecutive RS frames. Then they adopted a differentiable forward warping module to warp the RS image into the global one. Further considering the blurs in the RS distorted images, Zhong~\etal~\cite{zhong2021towards} proposed the first real-world rolling shutter correction and deblurring~(RSCD) and a joint correction and deblurring~(JCD) model to tackle the the RSCD problem. Most recently, Fan~\etal~\cite{fan2021sunet} utilized PWC-Net~\cite{sun2018pwc} to predict symmetric undistortion fields and restore the potential GS frames by a time-centered GS image decoder network, achieving promising results on the synthetic datasets. These methods still suffer from the blurs and detail loss in the restored GS frame due to the inaccurate displacement field estimation and warping. To alleviate such artifacts, we propose to predict multiple fields and warp the RS features adaptively.

\subsection{Attention Mechanism}
Attention mechanism was introduced~\cite{bahdanauCB2015neural} for machine translation, and has been widely used in both natural language processing and computer vision. In \cite{vaswani2017attention}, a novel Transformer architecture was constructed using attention as a primary mechanism, and it replaced the recurrent structure with the self-attention operation. Thanks to the powerful long-range and relation modeling capacities of attention, it was gradually introduced to vision tasks and has achieved considerable success~\cite{wang2018non, ramachandran2019stand, huang2019ccnet}.

Recently, attention mechanism or Transformer has been adapted to image or video restoration tasks and achieved great success, \eg, super resolution~\cite{yang2020learning, chen2021pre, liang2021swinir}. In~\cite{yang2020learning}, the authors proposed a texture transformer network for reference-based image super-resolution, which adopts an attention mechanism to transfer the texture details from the reference image adaptively. Chen~\etal~\cite{chen2021pre} proposed an Image Process Transformer (IPT) for various image restoration tasks, \eg, super-resolution, denoise, by task-specific heads and tails. Liang~\etal~\cite{liang2021swinir} utilized Swin Transformer~\cite{liu2021swin} for multiple image restoration tasks and achieved better performance with much fewer parameters. Attention has shown high potential for vision tasks. This paper also exploits the attention mechanism for adaptive warping to restore high-quality GS frames.

\subsection{RSC Dataset Synthesis}
\begin{figure}[t]
    \centering
    \includegraphics[width=\linewidth]{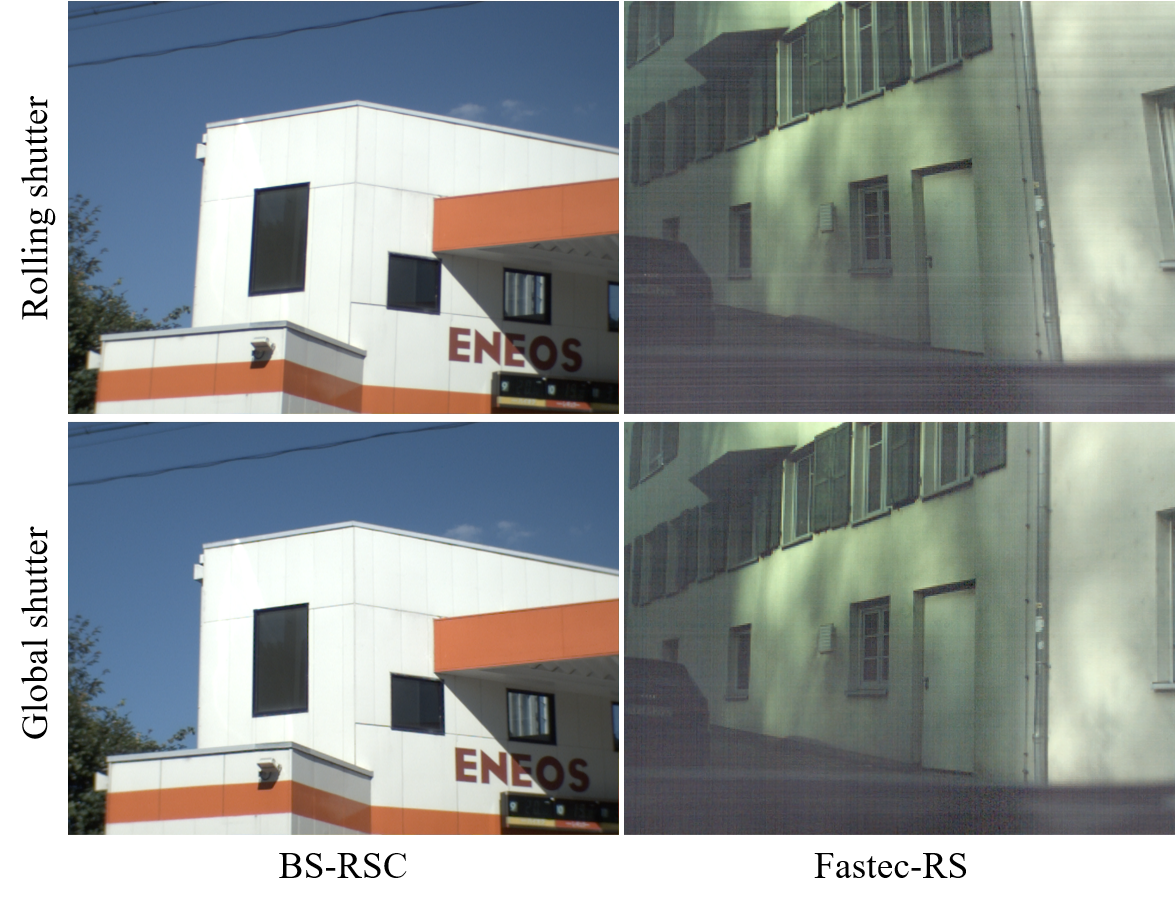}
    \vspace{-2em}
    \caption{\textbf{Left}: The real world RS-GS example in the collected BS-RSC dataset. \textbf{Right}: The synthesized RS-GS example in the Fastec-RS dataset~\cite{liu2020deep}. We see that our real RS frame is more natural, and there are much artifacts in the synthesized RS frames. }
    \label{fig:bs-rsc_example}
    \vspace{-4mm}
\end{figure}

Note that CNN-based approaches usually require a large amount of training data to learn the correction from RS to GS image. However, current RSC data or publicly available datasets are synthesized, where the RS images are generated from the captured GS images. For example, in ~\cite{rengarajan2017unrolling}, an affine transformation corresponding to RS motions is used to synthesize RS images. Zhuang~\etal~\cite{zhuang2019learning} synthesized RS images by warping a single GS image from KITTI dataset~\cite{geiger2012we} with dense depth map and camera motions. In~\cite{albl2020two}, various simulated motions are used to generate RS images. Recently, researchers in~\cite{liu2020deep} proposed two datasets, Carla-RS and Fastec-RS datasets, which generate more realistic RS distorted images via high-speed cameras and simulate the natural RS image formation process beyond camera motions or 3D geometry. The Carla-RS is synthesized from a free-moving rolling shutter camera in a virtual 3D Carla simulator.
On the contrary, the Fastec-RS dataset is created using the GS images in the real world with a 2400 FPS global shutter camera. However, the synthesized RS images are unnatural and full of line artifacts~(shown in~\cref{fig:bs-rsc_example}). Moreover, most of the scenes in Fastec-RS are collected by a horizontally moving camera, while various motions cause the RS images in the real world. These limitations significantly deteriorate the performance of RSC models. This paper proposes the first real-world RSC dataset for model training to restore high-quality GS images from real-world RS distorted images.

\begin{figure*}[htbp]
    \centering
    \includegraphics[width=0.99\linewidth]{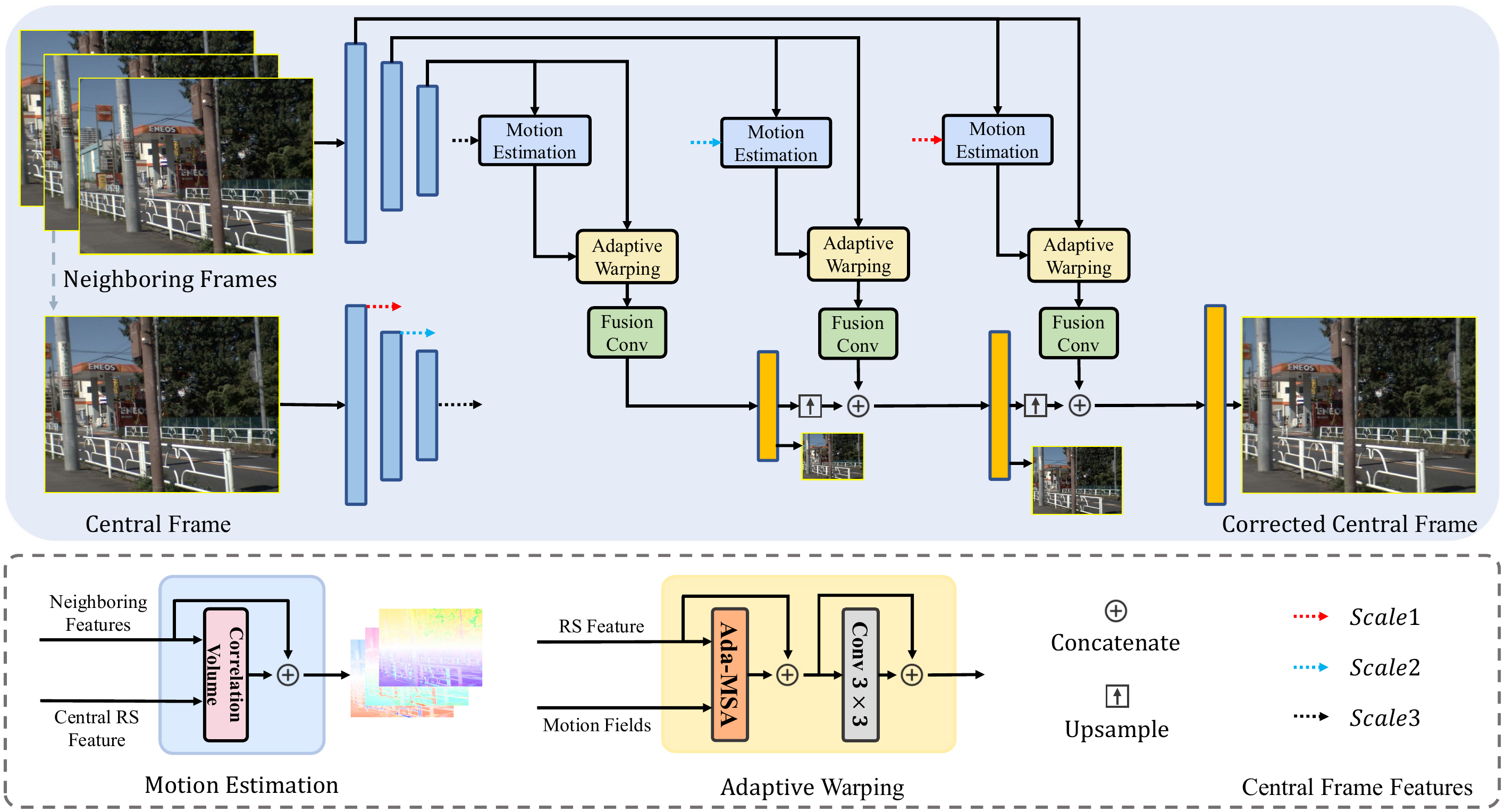}
    \caption{Main architecture of the proposed RSC model. Our model tries to predicts multiple displacement fields rather than only one to alleviate existing inaccurate motion estimation. We also propose an adaptive warping module to warp the RS features into the GS one adaptively under the guidance of the bundle of fields.}
    \label{fig:main_arch}
    \vspace{-5mm}
\end{figure*}
\section{Proposed Method}

\subsection{Problem Formulation}
As described in~\cite{liu2020deep}, the GS frame can be restored by warping the RS features backward with predicted displacement filed:
\begin{equation}
    \mathbf{I}^{g}(x) = \mathbf{I}^{r}(x+\mathbf{U}_{g\rightarrow r}(x)),
\end{equation}
where $\mathbf{I}^{g}$ is the potential GS frame; $\mathbf{I}^{r}$ is the input RS frame; $\mathbf{U}_{g \rightarrow r}$ is the displacement field from GS to RS frame, and $x$ is a certain pixel. It is difficult to estimate the displacement field $\mathbf{U}_{g \rightarrow r}$ since only the RS frames are available. Fortunately, the velocity vector can be estimated from the optical flow $\mathbf{V}$ between two consecutive RS frames. Thus the displacement can be calculated when the velocity is constant:
\begin{equation}
    \mathbf{U}(x) = \lambda \mathbf{V}(x) \mathbf{T}(x),
    \label{eq:displacement_velocity}
\end{equation}
where $\lambda$ is a scaling factor, and $\mathbf{T}(x)$ is the time offset corresponding to the middle scanline of RS frame. Therefore, existing methods try to estimate the displacement filed firstly from two consecutive RS frames, then warp the RS features with a differentiable forward warping block~(DFW)~\cite{liu2020deep}. A DFW module attempts to approximate the intensity of a particular pixel $x$ in the potential GS image by aggregating its neighboring pixel intensities in the RS features with weights proportional to the neighbor's distance, \ie, the greater the distance of a neighbor, the smaller its weight. Therefore, accurate motion estimation and warping are two key factors to restore the potential GS frames. However, the accurate $\mathbf{U}$ is hard to estimate since $\mathbf{U}$ cannot be effectively supervised during training when only the GS frame is adopted for supervision. The inaccurate estimation $\mathbf{U}$ further results in undesired warping results since the DFW module aggregates the neighboring pixels to $x$ with distance-aware weights. As a result, the corrected GS frame often suffers from blurs and other artifacts. 

\subsection{Model Overview}
Our model aims to alleviate inaccurate displacement field estimation and error-prone warping problems with multiple fields prediction and adaptive warping module. Building on current CNN-based RSC methods, our model inputs three consecutive RS frames to explore motion information and complementary contextual information, and restore the GS frame at the intermediate exposure time~(middle scanline) of the input central RS frame. Our model consists of three parts shown in~\cref{fig:main_arch}: a multi-scale feature extractor, an adaptive warping module, and a coarse-to-fine GS frame decoder. We first extract frame-level multi-scale features. Then, for the features at each scale, the neighboring RS features are used to predict the forward and backward motion information and warped by the proposed adaptive warping module. These warped features are fused by a convolution block. Last, the decoder decodes the warped features and outputs the corrected GS frame in a coarse-to-fine manner. 

\subsection{Adaptive Warping Module}
\noindent \textbf{Multiple Displacement Fields Generation. }
A key difference from previous methods is that our model predicts multiple displacement fields rather than one for warping. Moreover, the constant velocity assumption is too restrictive in~\cref{eq:displacement_velocity}, thus we modulated the multiple displacement fields by further predicting weights. Specifically, for the $t$-th RS feature $F^{l}_t \in \mathbb{R}^{C\times H \times W}$ at $l$-th scale, we first construct a 3D correlation volume $\mathbf{CV^{l}_{t}}$~\cite{sun2018pwc} to build the correspondence with central RS frames. Then the volume is used to predict multiple displacement fields and their weights by a residual block~\cite{he2016deep}:
\begin{equation}
    \{\mathbf{U}^{l, 0}_{t}, \dots, \mathbf{U}^{l, M-1}_{t}, \textbf{W}\} = \text{ResBlock}([\mathbf{CV}^{l}_{t}, F^{l}_t]),
\end{equation}
where $l$ index the scale, and $M$ is the number of motion fields. Each field contains two channels corresponding to the horizontal and vertical movements. $\textbf{W}\in \mathbb{R}^{M\times H \times W}$ is the weight of each estimated field. So the final predicted displacement fields are modulated by multiplying the estimated weights.

\begin{figure}[!tbp]
    \centering
    \includegraphics[width=\linewidth]{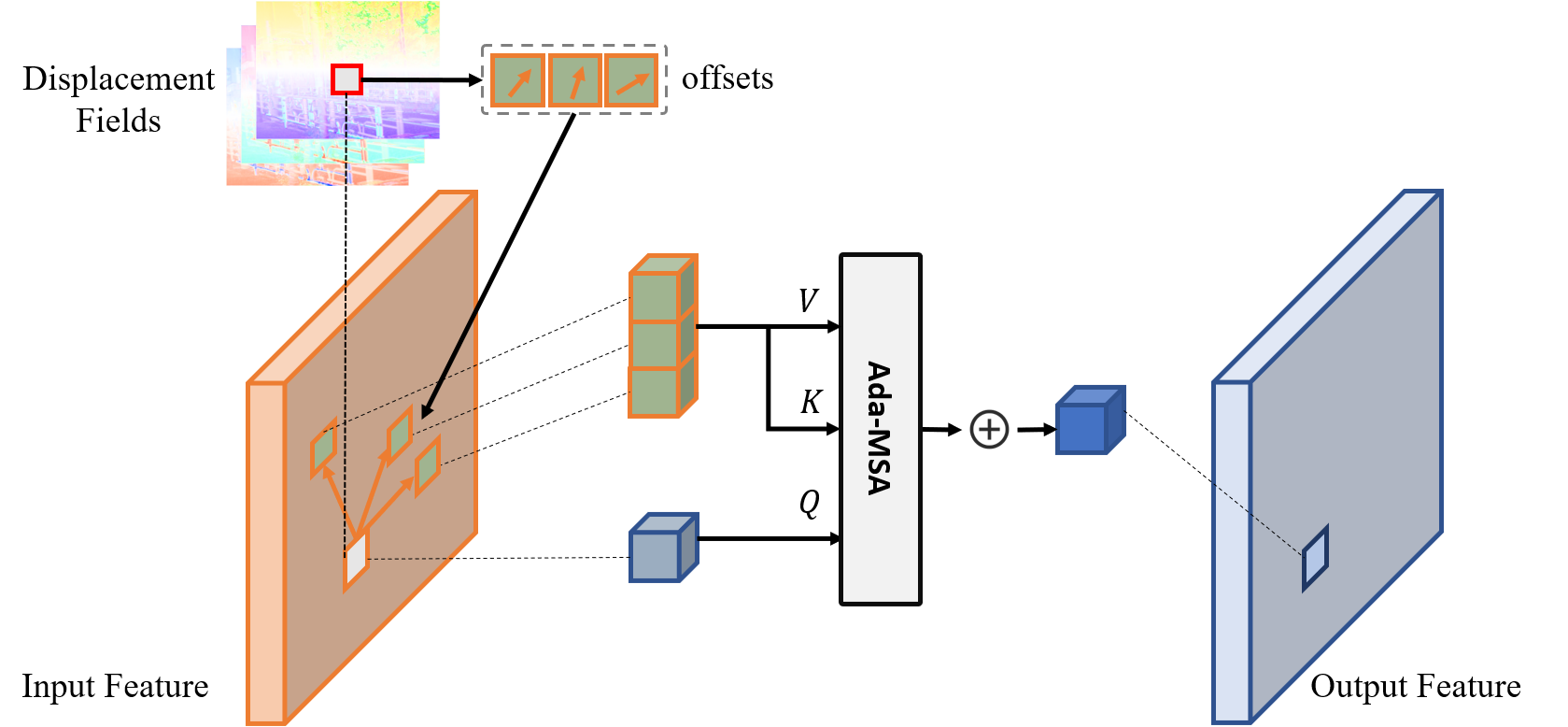}
    \caption{Illustration of adaptive multi-head self-attention mechanism~(Ada-MSA). Ada-MSA aims to warp the input RS features into the GS features adaptively under the guidance of estimated multiple motion fields.}
    \label{fig:ada_msa}
    \vspace{-2mm}
\end{figure}

\vspace{2mm}
\noindent \textbf{Adaptive Warping.}
As for the warping process, we proposed an Adaptive Warping Module~(AWM) utilizing self-attention to aggregate the features sampled under the predicted multiple displacement fields. AWM consists of an adaptive multi-head attention~(Ada-MSA) and a convolutional block. The Ada-MSA mechanism is shown in \cref{fig:ada_msa}. Firstly, for each pixel $x$~(consists of row index $i$ and column index $j$) in $t$-th RS features $F^{l}_t$ at scale $l$, the query vector $Q$ is generated by a linear transformation with matrix $W_q$:
\vspace{-1mm}
\begin{equation}
    Q = W_q F^t_{l}(x).
\end{equation}
Subsequently, the feature set $N(x)$ are sampled under the guidance of estimated multiple displacement fields $\mathbf{U}^{l}_{t}$:
\vspace{-1mm}
\begin{equation}
    N^{l}_{t}(x)=\{F^{l}_{t}(x + \mathbf{U}^{l, i}_{t}(x))|i=0, 1, \dots, M-1\}.
\end{equation}
Then the key $K$ and value $V$ vectors are then generated by a linear transformation from the sampled features:
\begin{equation}
    K = W_k N^{l}_{t}(x), V = W_v N^{l}_{t}(x),
\end{equation}
where $W_k \in \mathbb{R}^{d \times C}$ and $W_v \in \mathbb{R}^{d \times C}$ are transformation matrices. Thus the adaptive attention feature at $h$-th head is calculated by
\begin{equation}
    \text{AdaMSA}_h(x, F^{l}_{t}, \mathbf{U}^{l}_{t}) = \text{SoftMax}(\frac{Q_{h}^{T}K_{h}}{\sqrt{d_h}})V_{h}^T,
\end{equation}
where $h$ indexes the attention head, and $Q_{h}$, $K_{h}$ and $V_{h}$ are with dim $d_h=\frac{d}{H}$. The outputs of all $H$ heads are concatenated into $d$ dims vector and projected to the output feature. Through this adaptive warping module with multiple multiple motion fields, the RS features are aggregated to the GS counterpart adaptively.

\subsection{Loss Functions}
We train the proposed model in an end-to-end manner, and only the ground truth GS frame is required for supervision. Following previous work~\cite{zhong2021towards}, we adopt the Charbonnier loss $\mathcal{L}_c$ and perceptual loss $\mathcal{L}_p$ to ensure the visual quality of the corrected GS frame. The total variation loss $\mathcal{L}_{tv}$ is adopted to ensure the smoothness in the estimated displacement field. Thus total loss can be formulated as:
\begin{equation}
    \mathcal{L} = \mathcal{L}_c + \lambda_p\mathcal{L}_p + \lambda_{tv}\mathcal{L}_{tv}.
\end{equation}

\section{BS-RSC Dataset}
\label{sec:rsc_dataset}
A real-world dataset without synthetic artifacts is essential to improve the capacity of real applications of CNN-based RSC methods. Recently, some specific optical acquisition systems have been designed to capture the real-world image or video pairs for restoration tasks, improving the generalization capacity of CNN models. Cai~\etal~\cite{cai2019toward} constructed a real-world super-resolution dataset where paired high- and low-resolution data of the same scene are captured by adjusting the focal length of a digital camera. For deblurring, Rim~\etal~\cite{rim2020real} and Zhong~\etal~\cite{zhong2020efficient} collected real-world single image and video deblurring dataset respectively, adopting a beam-splitter acquisition system. Inspired by these pioneering works, we also propose a beam-splitter acquisition system to collect the first real-world dataset for the RSC task, termed as BS-RSC. 

\begin{figure}[b]
    \centering
    \includegraphics[width=\linewidth]{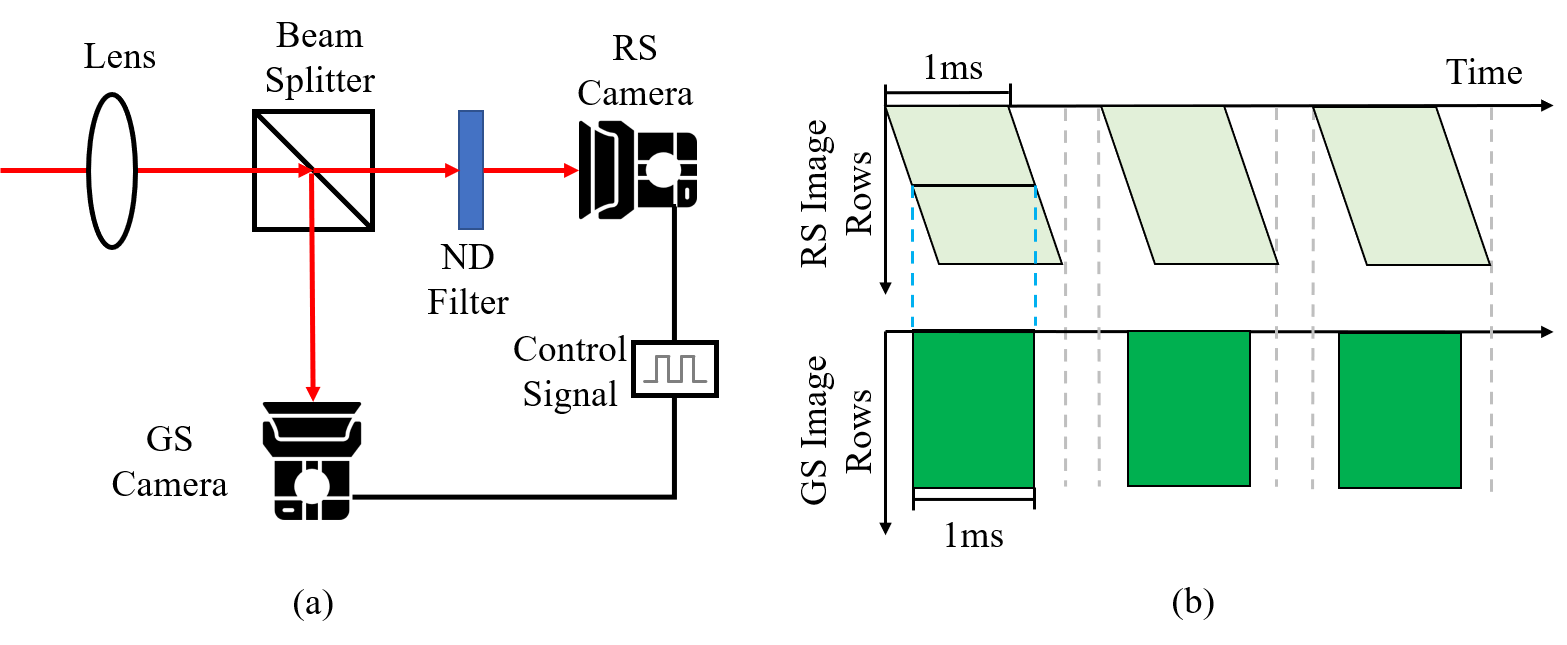}
    \caption{The designed beam-splitter acquisition system for real-world RSC dataset construction. \textbf{(a)} structure of the designed beam-splitter acquisition system. \textbf{(b)} exposure scheme of the GS and RS camera. The acquisition system can capture the GS frame at the intermediate exposure time of RS frame. }
    \label{fig:bs_acquisition_system}
\end{figure}

\vspace{-1mm}
\subsection{Beam-Splitter Acquisition System}

The architecture of the designed beam-splitter acquisition system is shown in~\cref{fig:bs_acquisition_system}(a), where a beam-splitter splits the incoming light into two beams and passes them into the following RS and GS cameras. We choose the FLIR FL3-U3-13S2C RS camera with a 1/3-inch CMOS sensor (3.63 um pitch size) and the FLIR GS3-U3-28S4C GS camera with a 1/1.8-inch CCD sensor (3.69 µm pitch size). These two cameras are geometrically aligned via the 50/50 beam splitter. With the aid of a laser beam, we first adjust the alignment mechanically towards an accuracy of a few pixels. After that, we conduct a homography correction with a standard checker pattern to further reduce misalignment to subpixel level. The exposure time of both the RS and the GS camera is 1ms, avoiding blurs in the captured video. Both cameras run at 25 fps. We use a wave generator to generate synchronized pulses at 25Hz, and the phase of the pulse for the GS camera is properly delayed, such that the GS exposure timestamp matches the middle scanline of the RS camera (shown in~\cref{fig:bs_acquisition_system}(b)). As for photometric alignment, we put a neutral density filter before the RS camera to equalize the sensitivity of the two cameras. We further use a color checker pattern to correct the RGB response of the GS camera, such that both cameras share the same color response. The whole system is just about one kilogram, thus can be held easily and moved freely. 

\vspace{-1mm}
\subsection{Data Composition}
The collected BS-RSC contains RS videos with various camera and object motions, mainly in outdoor street scenes with cars and people, \etc. Specifically, the designed beam-splitter acquisition system collects a total of 80 RS-GS HD~($1024 \times 768$) video pairs, and each video contains 50 frames. We further divide it into Train set, Val set, and Test set with 50, 15, 15 videos, respectively.

\section{Experiments}
\subsection{Experimental Setting}
\noindent \textbf{\normalsize Datasets.}
We conduct experiments on the proposed real-world BS-RSC dataset. Besides, we also provide experimental results on the popular synthesized dataset Fastec-RS~\cite{liu2020deep}, which contains 76 video sequences, and each video contains 34 frames. Note that we use the dataset coming from the public released dataset, which is slightly different from the description in the original paper, and both the test and validation subsets are used to calculate the metrics.

\vspace{2mm}
\noindent \textbf{Implementation Details.}
During training, three consecutive RS frames in RGB style are fed into our model. The input frames are first randomly cropped into $480\times 256$ and randomly flipped horizontally for data augmentation. $\lambda_p$ and $\lambda_tv$ are set to $0.01$ and $0.001$, respectively. The initial learning rate is set to $2 \times 10^{-4}$, and the ADAM~\cite{kingma2014adam} is adopted to optimize the model parameters. The model is trained for 400 epochs with a cosine annealing learning rate adjusting scheduler. For testing, three consecutive frames are fed into the model directly without any augmentation. We set the number of displacement fields $M=9$ in the following experiments.

\vspace{2mm}
\noindent \textbf{Evaluation Metrics and Methods of Comparison.}
Both PSNR and SSIM are employed to evaluate the corrected results quantitatively. Visualizations of the corrected RS frames are shown for qualitative comparison. We compare the proposed method to state-of-the-art RSC method, including a traditional methods proposed in~\cite{zhuang2017rolling}, CNN-based methods DSUN~\cite{liu2020deep}, JCD~\cite{zhong2021towards} and SUNet~\cite{fan2021sunet}. These methods have shown promising effectiveness on the synthesized dataset. As the authors of SUNet have not yet published the code and test results, we cannot report any results other than those in the original paper.

\subsection{Comparison to the State-of-the-art.}
\begin{figure*}[!ht]\footnotesize
    \centering
    \begin{tabular}{@{\hspace{1mm}}c@{\hspace{0.5mm}}c@{\hspace{0.5mm}}c@{\hspace{1mm}}}
        \includegraphics[width=0.33\linewidth]{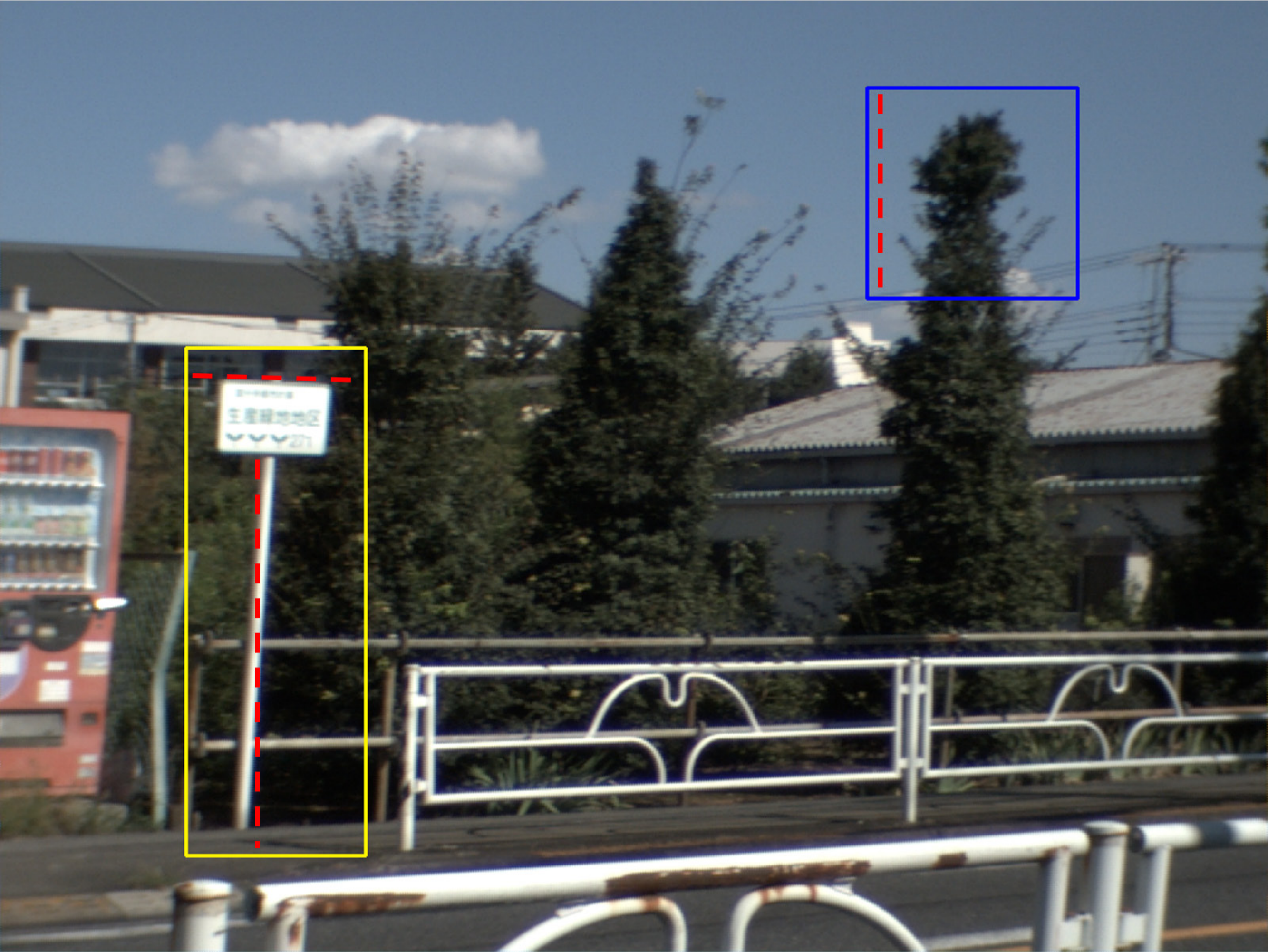} &
        \includegraphics[width=0.33\linewidth]{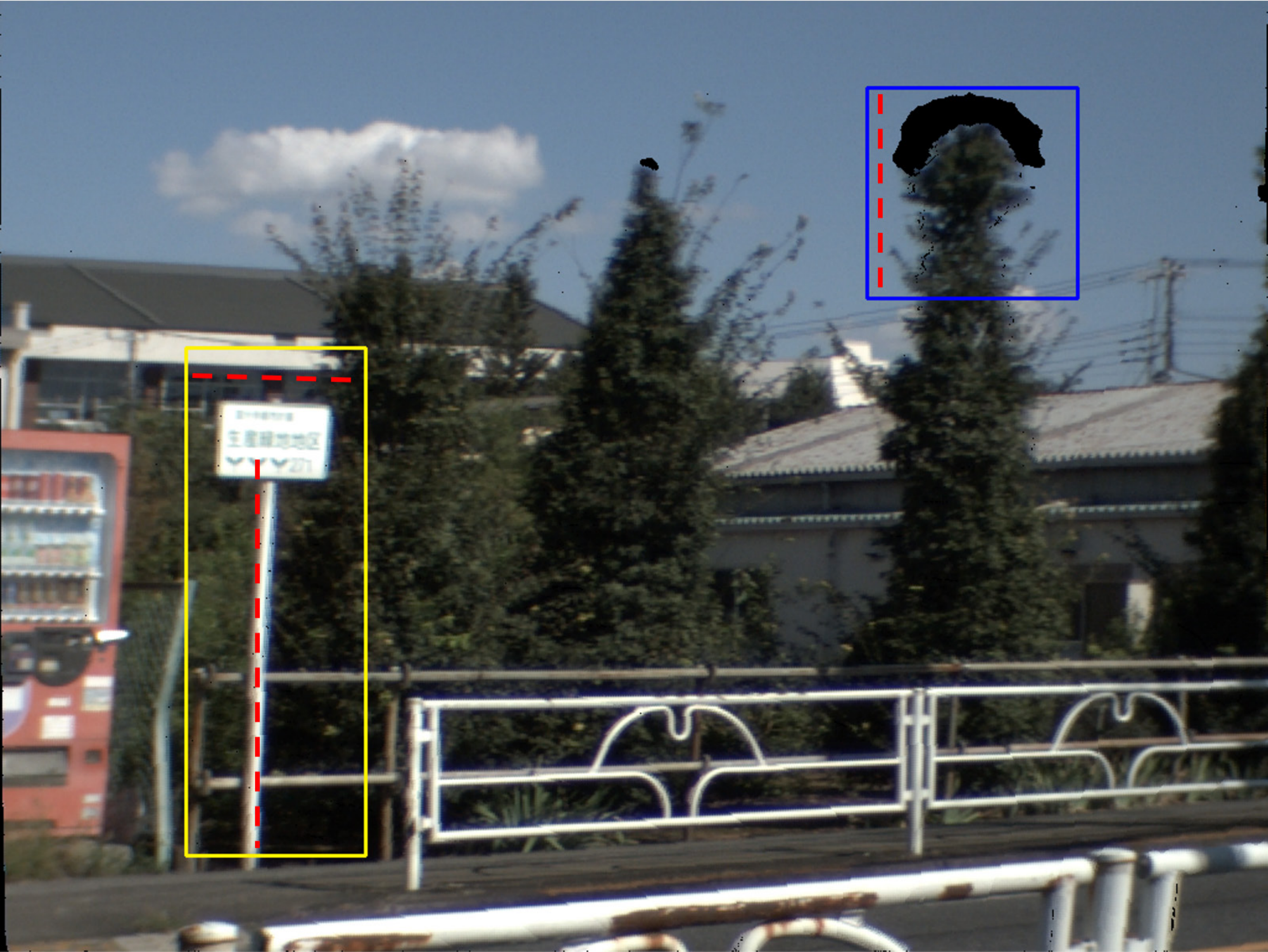} &
        \includegraphics[width=0.33\linewidth]{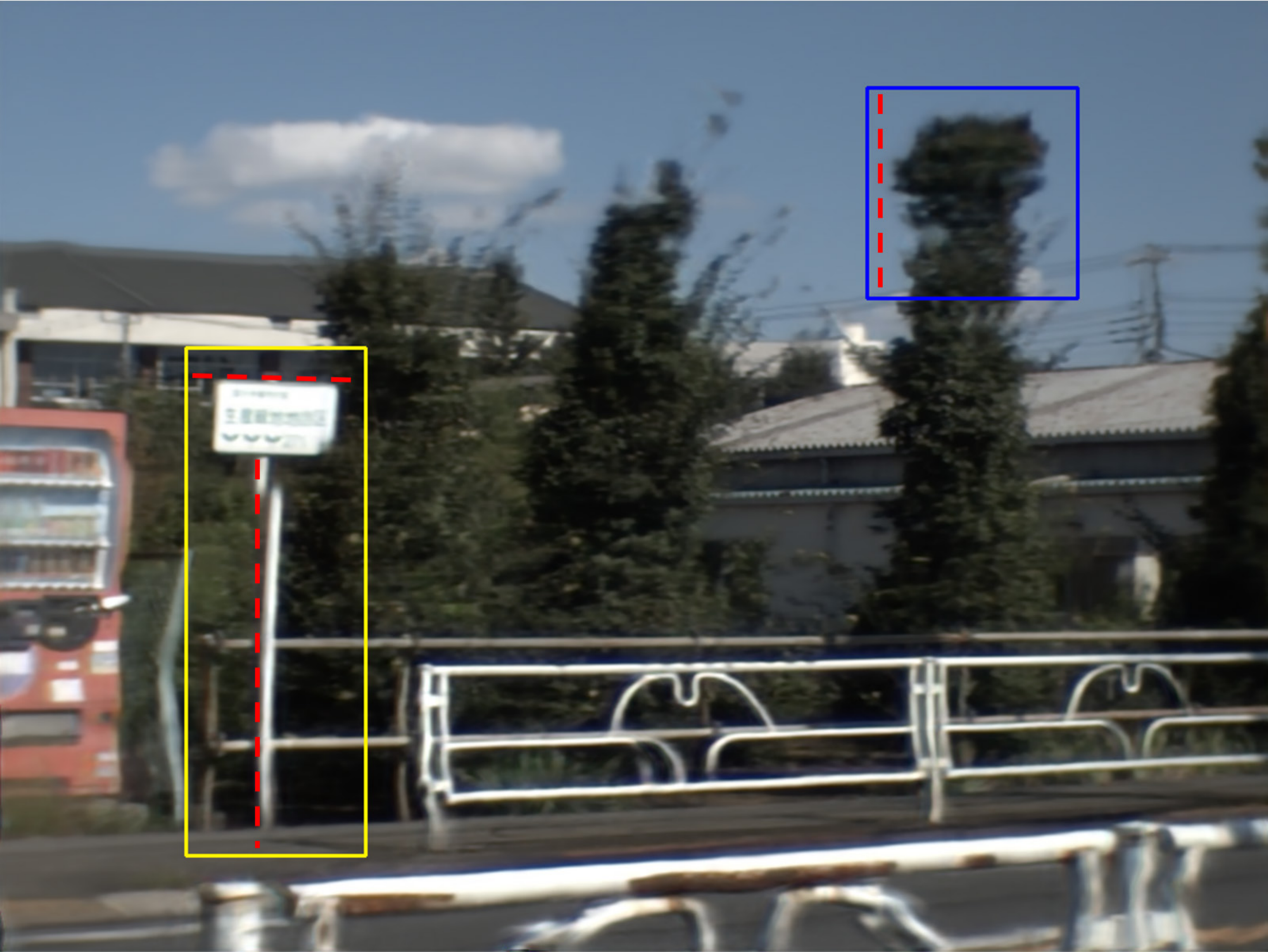}  \\
        (a) RS frame  & (b) Zhuang~\etal~\cite{zhuang2017rolling} & (c) DSUN~\cite{liu2020deep} \\
        \includegraphics[width=0.33\linewidth]{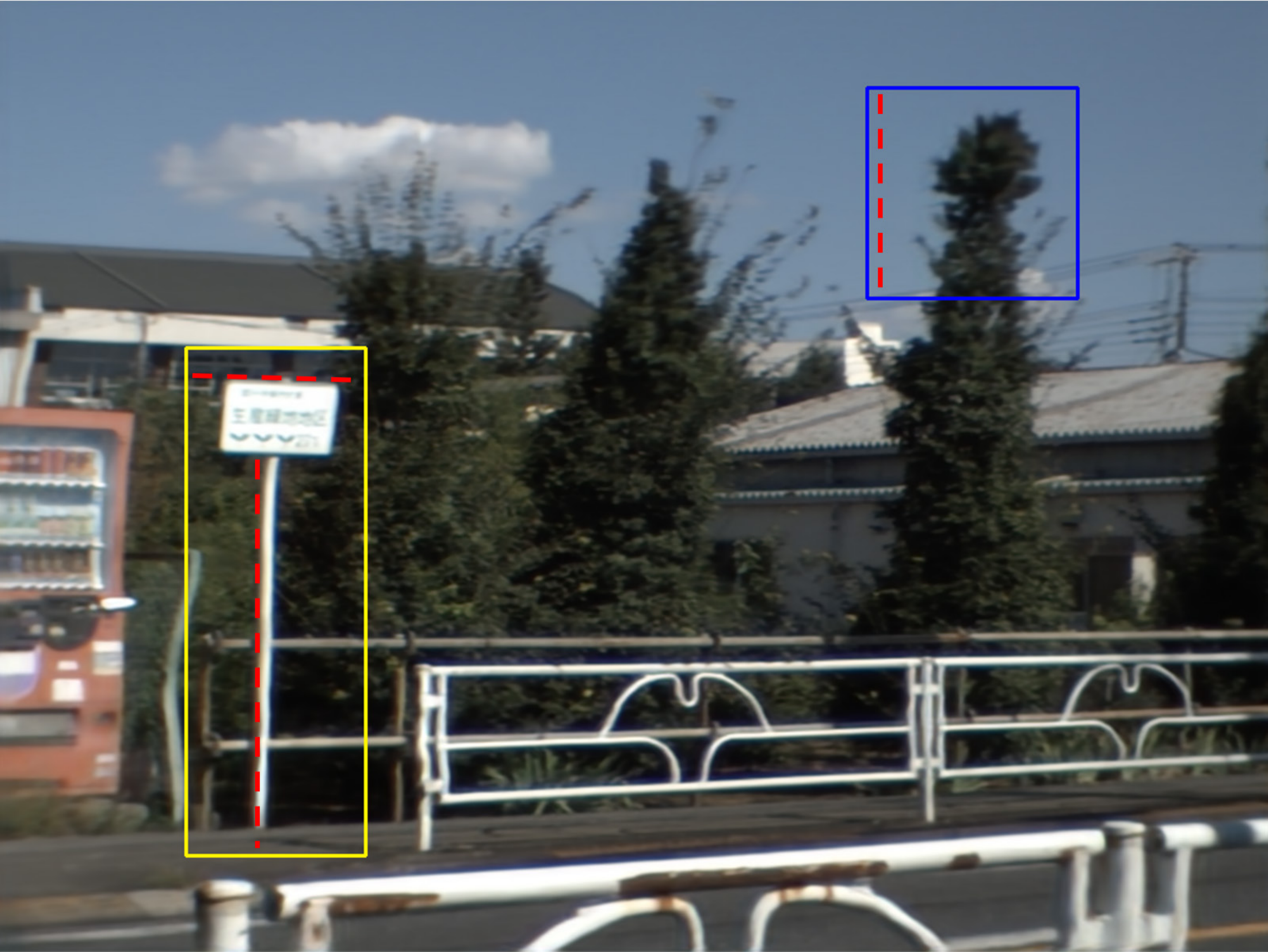} &
        \includegraphics[width=0.33\linewidth]{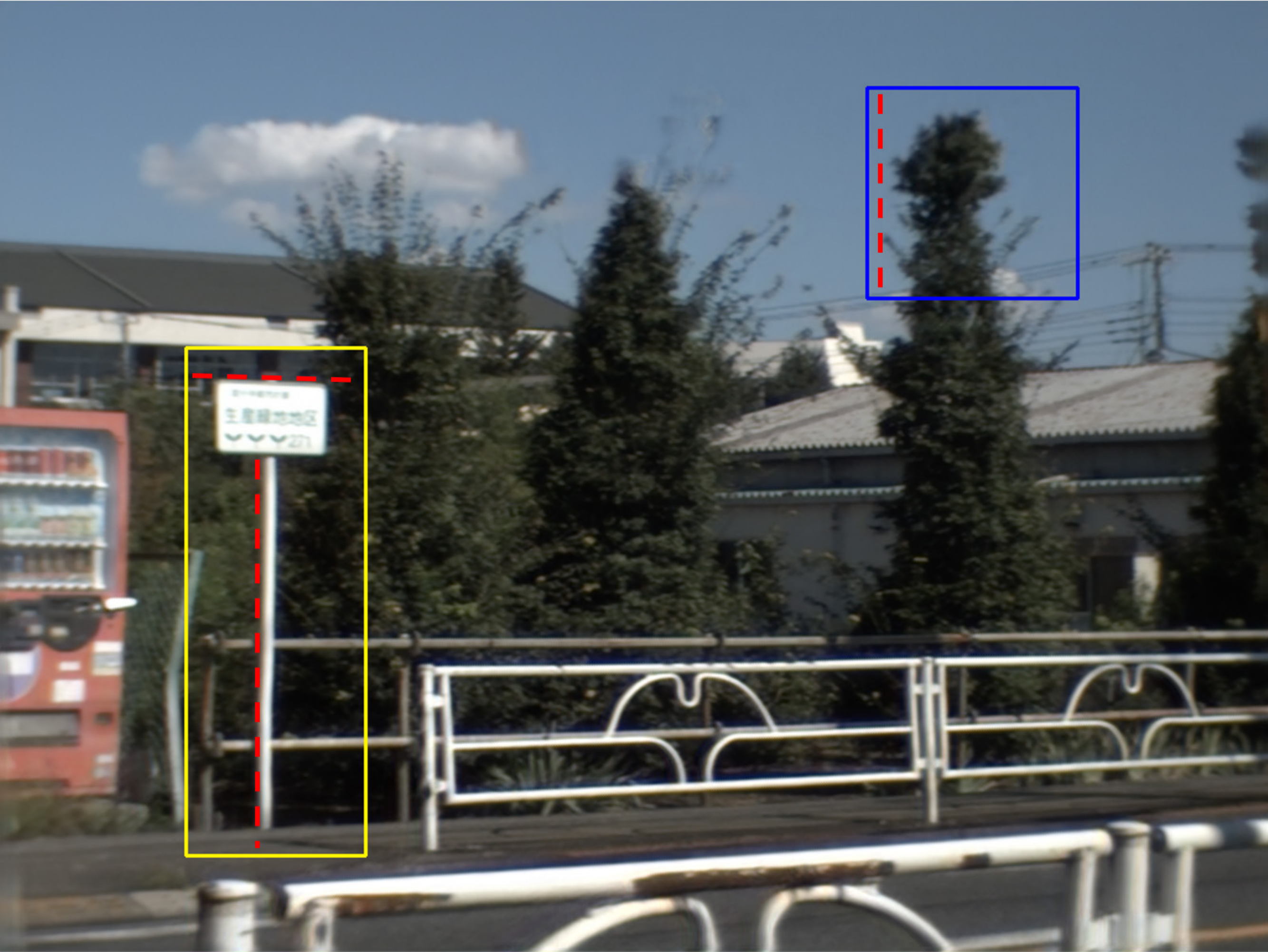} &
        \includegraphics[width=0.33\linewidth]{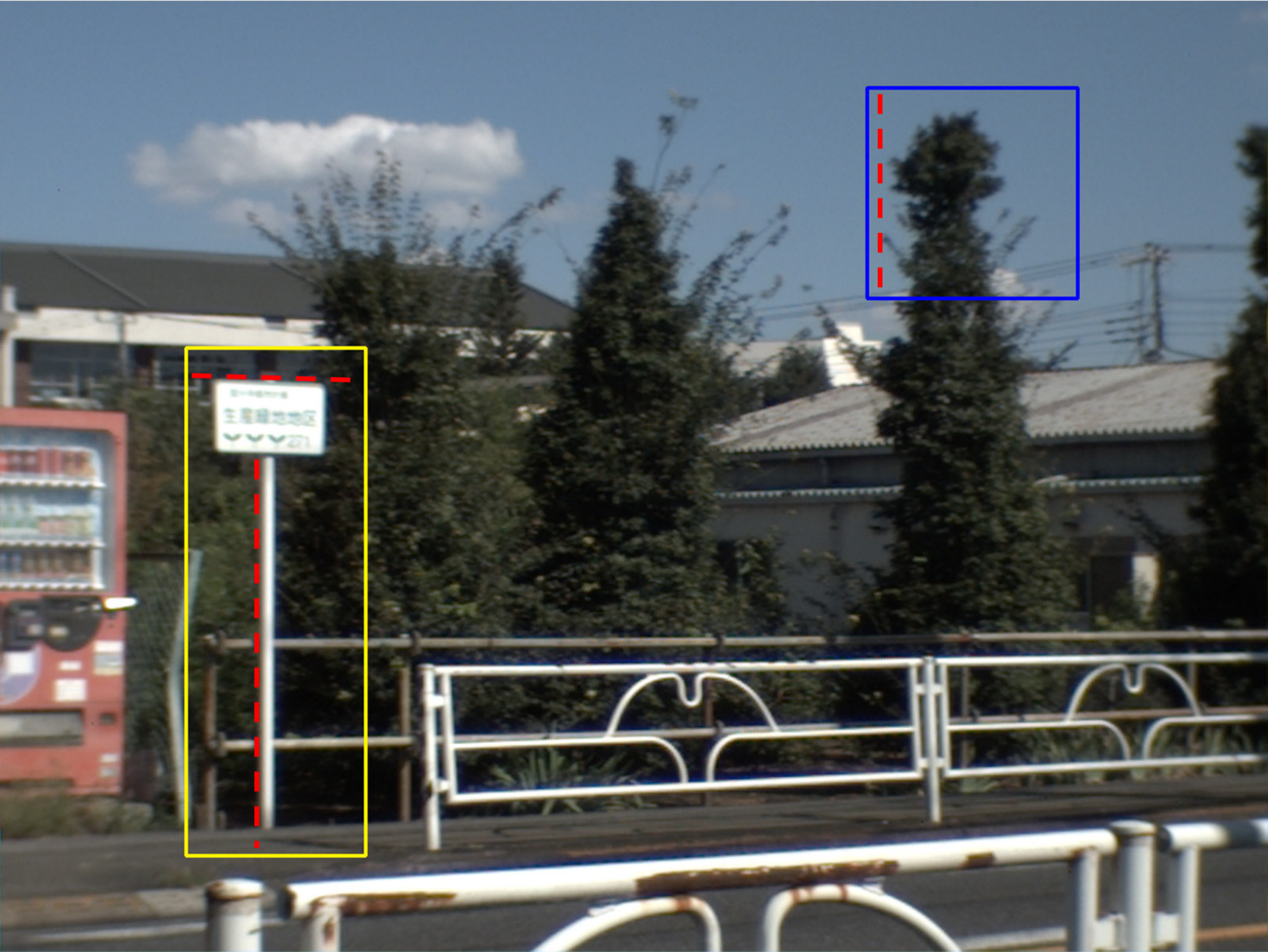}  \\
        (d) JCD~\cite{zhong2021towards} & (e) Ours & (f) GS frame \\
        \end{tabular}
        \caption{Visual comparison on the proposed BS-RSC dataset for real-world RSC. Our method obtains higher visual quality, and more details are restored with fewer artifacts. Though the existing methods obtained highly competitive results on the synthesized dataset, they failed to restore the real-world RS distortions due to the difficulty of modeling the challenging motion in the BS-RSC. } %
    \label{fig:restults_on_bsrsc}
\end{figure*}

\paragraph{Results on BS-RSC.}
The quantitative comparison of the proposed real-world dataset BS-RSC is shown in \cref{tab:resutls_on_bsrsc}. Thanks to the multiple motion fields prediction and the adaptive warping strategy, our model achieves the best PSNR and SSIM evaluation metrics with a large performance improvement than SOTA methods. The qualitative comparison is shown in~\cref{fig:restults_on_bsrsc}. We see that the proposed method obtains more visually friendly results than other methods ~(\eg, the billboard and the trees). These superior performances significantly demonstrate the effectiveness of our model on real-world rolling shutter correction.

\begin{table}[!htbp]
    \centering
    \begin{tabular}{l@{\extracolsep{0.5cm}}cc}
    \toprule
    Methods     &  PSNR$\uparrow$(dB)  &  SSIM$\uparrow$  \\
    \midrule
    Zhuang~\etal~\cite{zhuang2019learning}
                & 19.80  & 0.698 \\
    DeepUnrollNet~\cite{liu2020deep}
                & 23.60  & 0.808  \\
    JCD~\cite{zhong2021towards}
                & 24.86  & 0.820  \\
    \midrule
    Ours        & \textbf{28.23}  & \textbf{0.882} \\
    \bottomrule
    \end{tabular}
    \caption{Quantitative comparison against the state-of-the-art RSC methods on the proposed BS-RSC dataset. }
    \label{tab:resutls_on_bsrsc}
\end{table}

\begin{figure*}[!ht]\footnotesize
    \vspace{4mm}
    \centering
    \begin{tabular}{@{\hspace{1mm}}c@{\hspace{0.5mm}}c@{\hspace{0.5mm}}c@{\hspace{1mm}}}
        \includegraphics[width=0.33\linewidth]{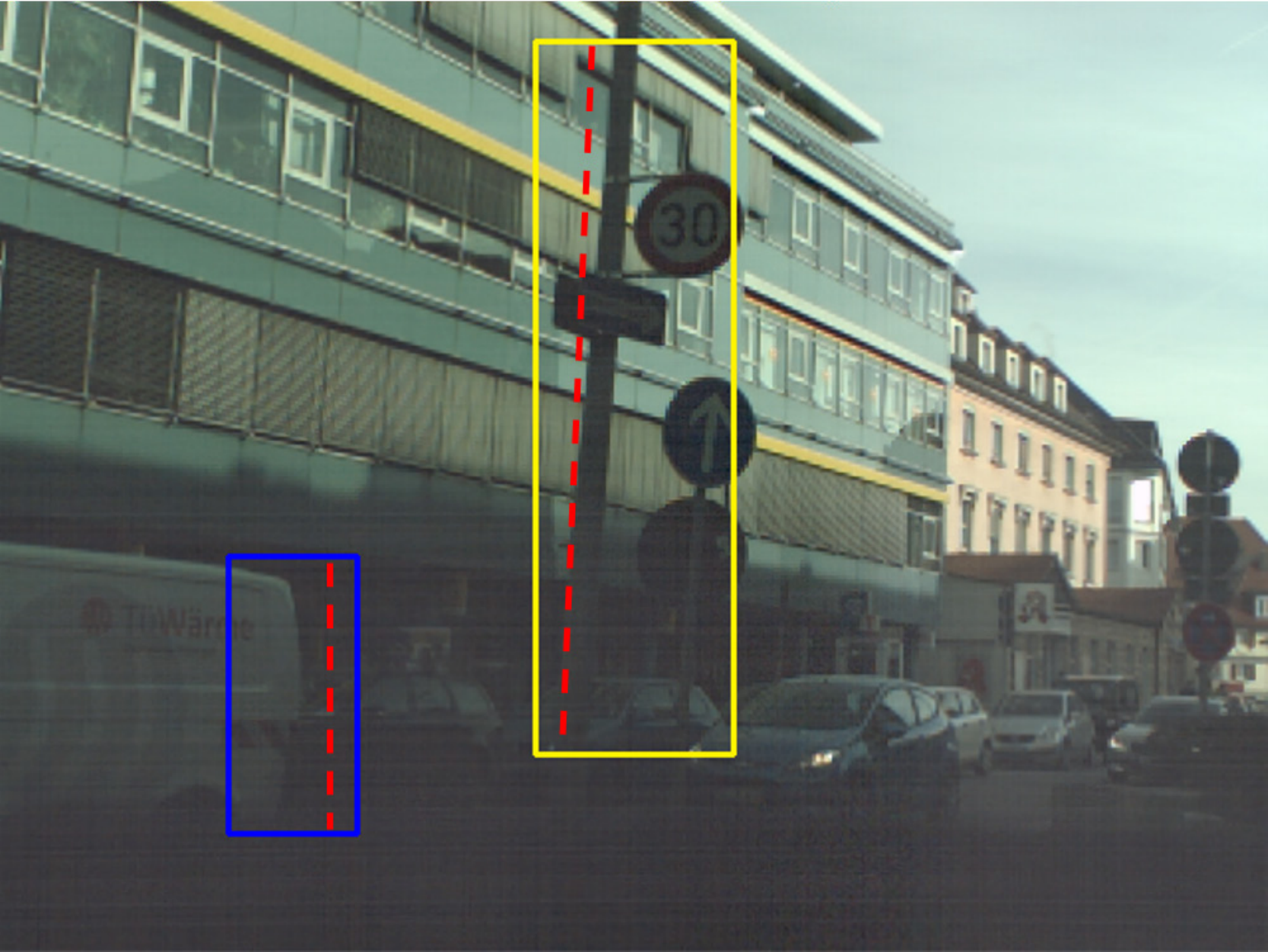} &
        \includegraphics[width=0.33\linewidth]{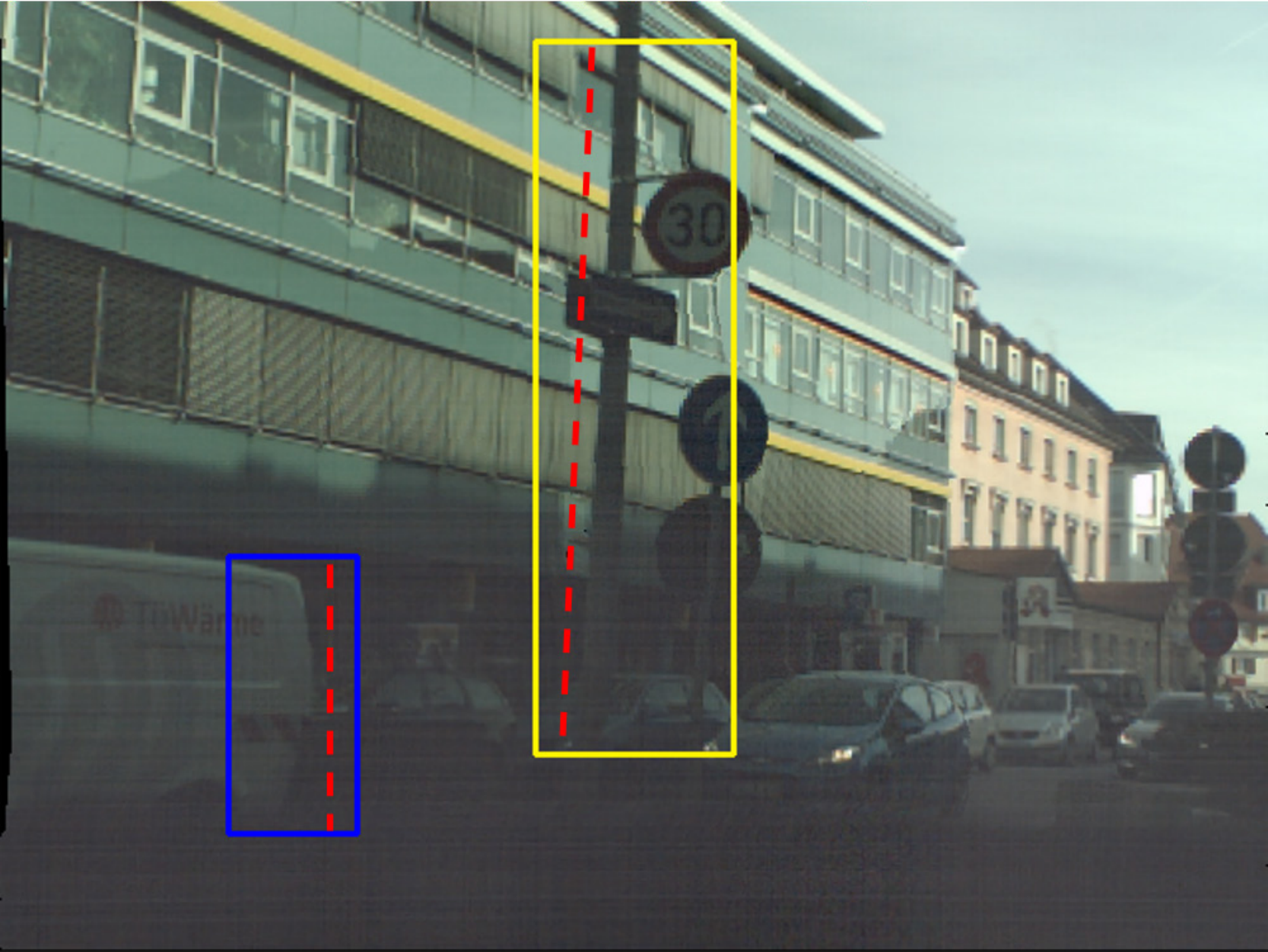} &
        \includegraphics[width=0.33\linewidth]{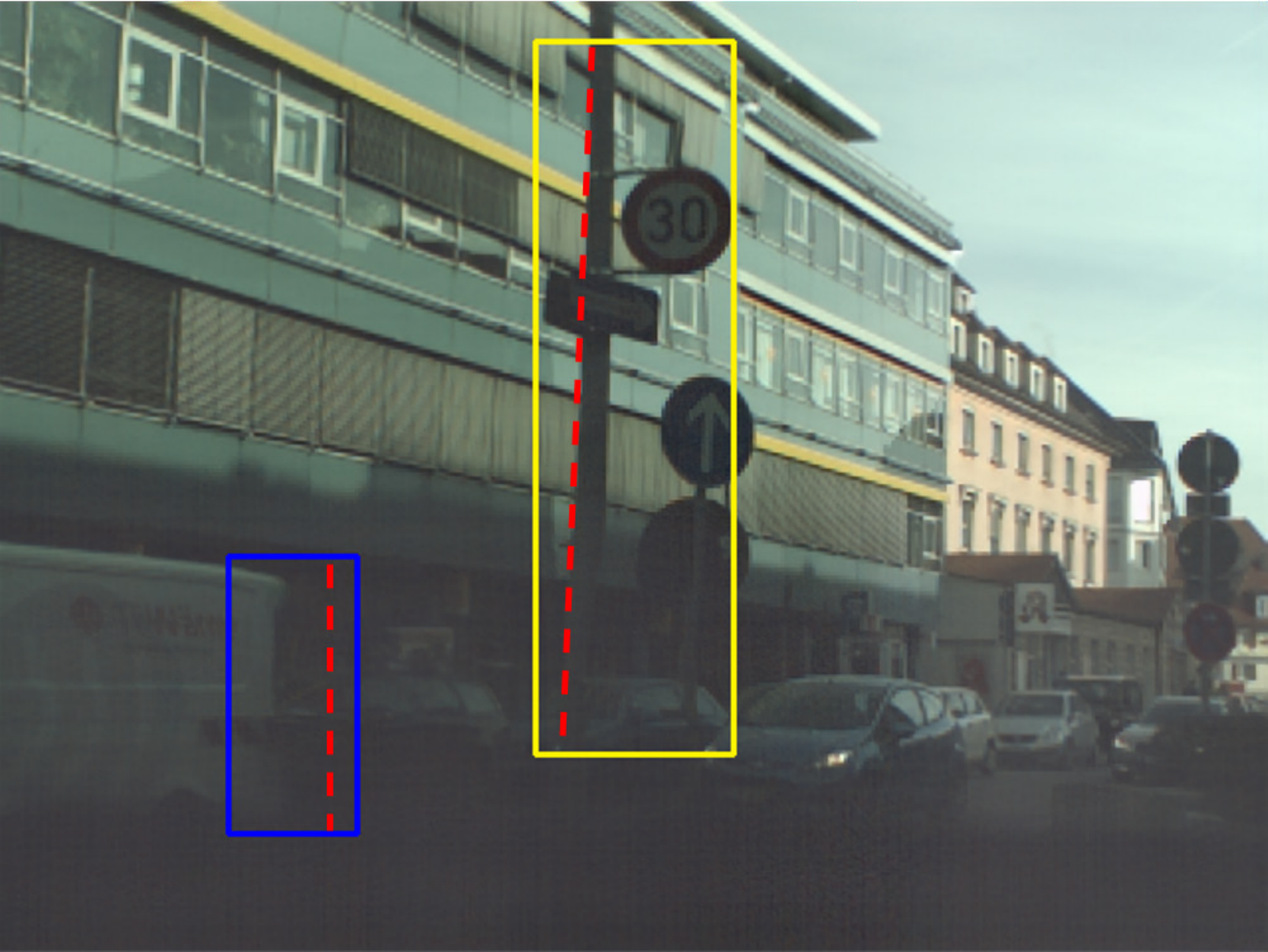}  \\
        (a) RS frame  & (b) Zhuang~\etal~\cite{zhuang2017rolling} & (c) DSUN~\cite{liu2020deep} \\
        \includegraphics[width=0.33\linewidth]{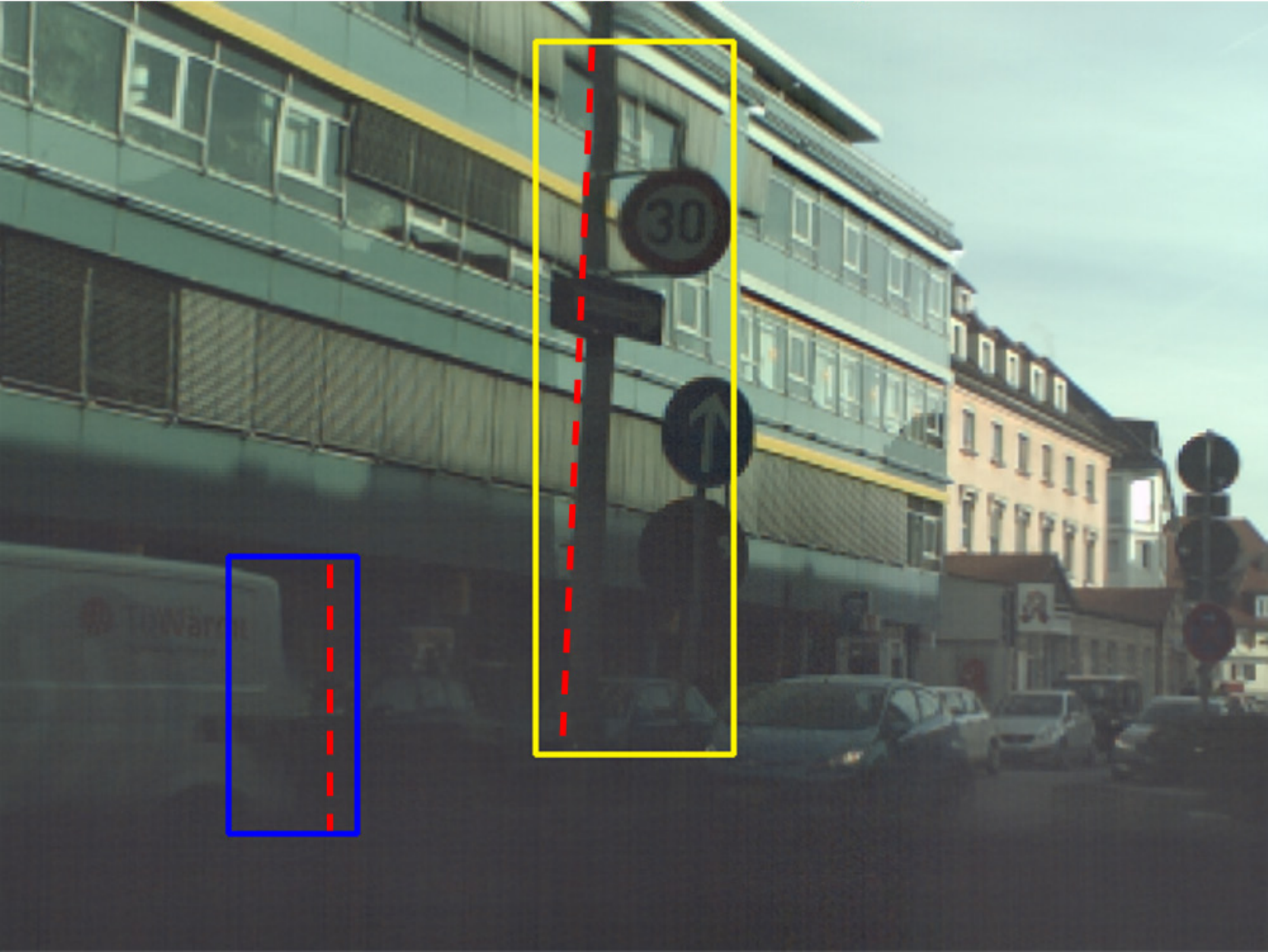} &
        \includegraphics[width=0.33\linewidth]{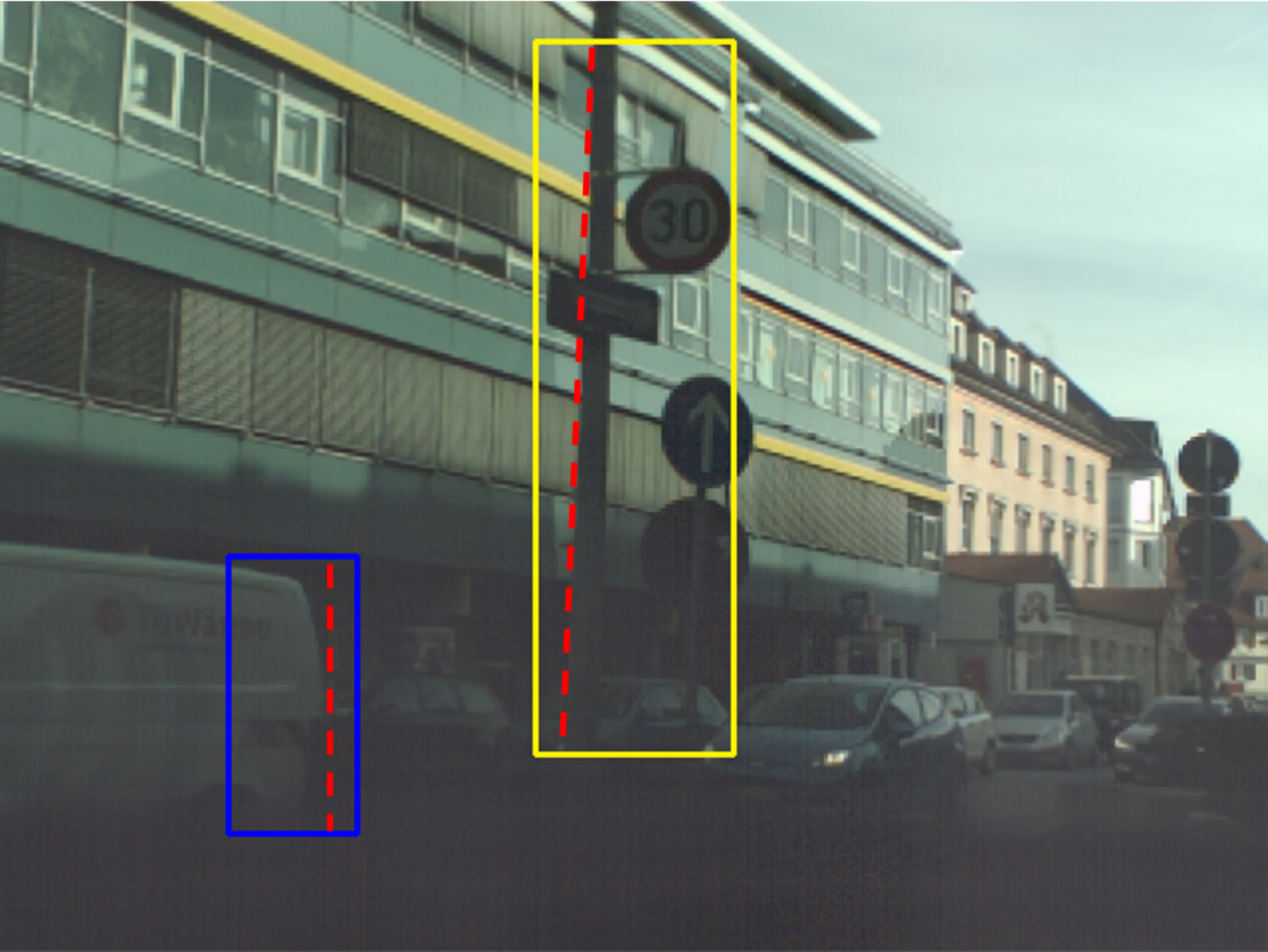} &
        \includegraphics[width=0.33\linewidth]{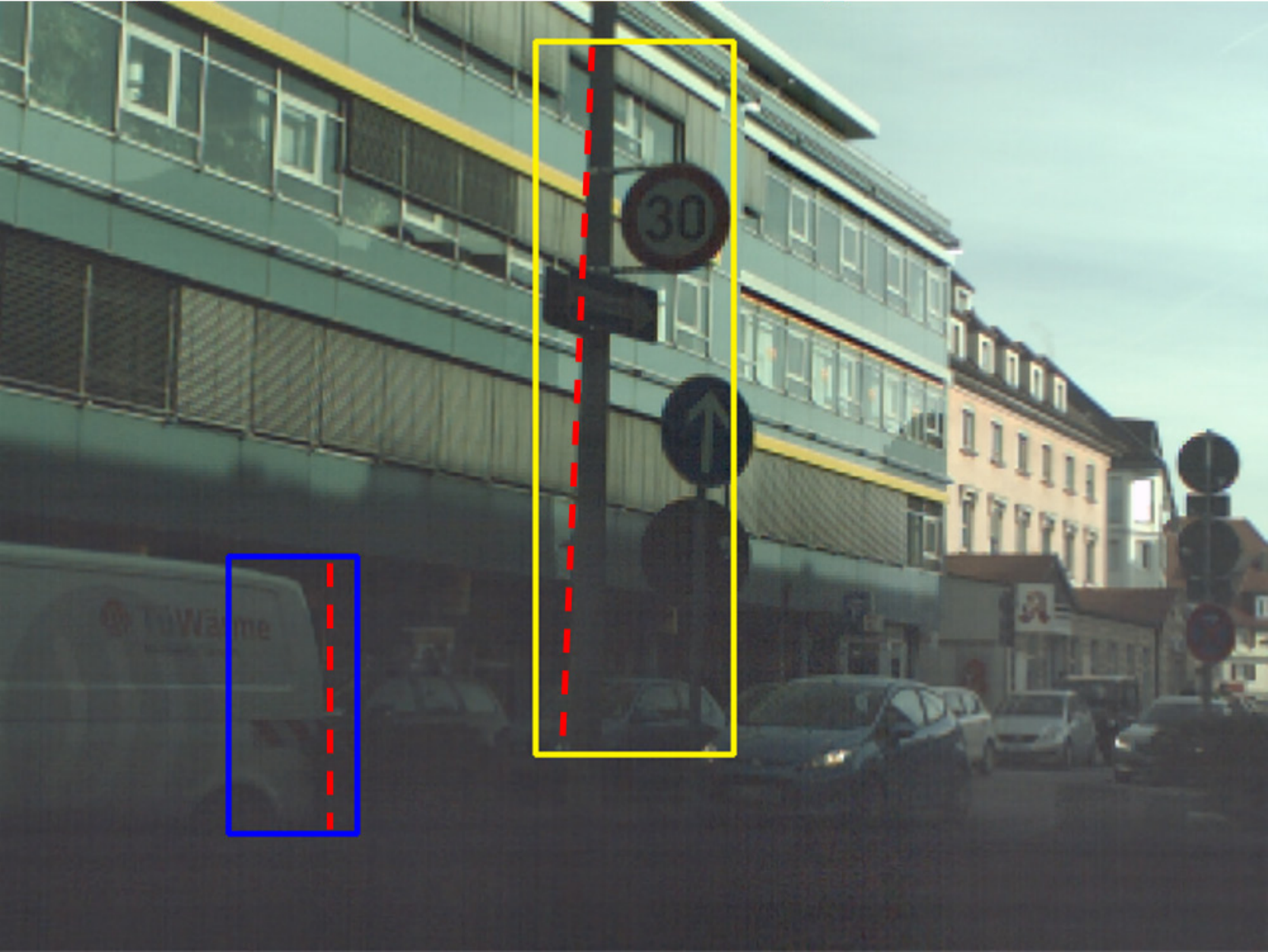}  \\
        (d) JCD~\cite{zhong2021towards} & (e) Ours & (f) GS frame \\
        \end{tabular}
        \caption{Visual results on the synthesized Fastec-RS dataset. The proposed method shows a strong competitive edge against other methods. } %
    \label{fig:restults_on_fastecrs}
\end{figure*}

\vspace{-2mm}
\noindent \textbf{Results on Fastec-RS.}
Besides the compassion on BS-RSC, we further evaluate the proposed method on the synthesized RSC dataset Fastec-RS to verify its effectiveness. The quantitative and qualitative results are shown in~\cref{tab:resutls_on_fastec} and \cref{fig:restults_on_fastecrs}, respectively. We see that our model achieves comparable evaluation results against other methods. 

These quantitative and qualitative results shown above demonstrate the superior performance of our model.

\begin{table}[!h]
    \centering
    \begin{tabular}{l@{\extracolsep{0.5cm}}cc}
    \toprule
    Methods     &  PSNR$\uparrow$(dB)  &  SSIM$\uparrow$ \\
    \midrule
    Zhuang~\etal~\cite{zhuang2019learning}
                & 21.44  & 0.710 \\
    DeepUnrollNet~\cite{liu2020deep}
                & 27.00  & 0.825  \\
    JCD~\cite{zhong2021towards}
                & 24.84  & 0.778 \\
    SUNet~\cite{fan2021sunet}*
                & 28.34  & 0.840 \\
    \midrule
    Ours        & \textbf{28.56}  & \textbf{0.855} \\
    \bottomrule
    \end{tabular}
    \caption{Quantitative comparison against the state-of-the-art RSC methods on the synthesized Fastec-RC dataset. * means that SUNet restores the GS frame at the first row of the RS frame.}
    \label{tab:resutls_on_fastec}
\end{table}

\vspace{-4mm}
\subsection{Ablation Study}
\noindent \textbf{Number of Input Frames. }
Our model takes three frames to  model the motion information more accurately for warping. To verify this, we modify our model to adapt single frame and two frames input. \cref{tab:ab_num_frames} presents the quantitative results with different number of inputs. A single frame input achieves the lowest metrics due to the ill-posed nature of estimating the displacement filed from a single frame. Instead, multi-frames can provide inter-frame movements and complementary information to perform better, especially when inputting the three consecutive RS frames. 
\begin{table}[!htbp]
    \centering
    \begin{tabular}{c@{\extracolsep{0.5cm}}cc}
    \toprule
    Input Frames&  PSNR$\uparrow$(dB)  &  SSIM$\uparrow$  \\
    \midrule
       1       & 23.84  & 0.765 \\
       2       & 27.20  & 0.838 \\
       3       & \textbf{28.56}  & \textbf{0.855} \\
    \bottomrule
    \end{tabular}
    \caption{Ablation study of the number of the input RS frames.}
    \label{tab:ab_num_frames}
    \vspace{-3mm}
\end{table}

\noindent \textbf{Adaptive Warping Module. }
To verify the effectiveness of the proposed warping module, we further construct three models. \textit{Net1} only adopts a convolution for multiple RS features fusion without any warping. \textit{Net2} replaces the AWM with the common backward warping. \textit{Net3} replaces the AWM with the DFW adopted by existing methods. The results shown in~\cref{tab:ab_awm} demonstrate the effectiveness of the proposed adaptive warping module. Meanwhile, our model achieves best PSNR and SSIM when number of the predicted motion fields $M$ equals 9.

\begin{table}[!htbp]
    \centering
    \begin{tabular}{c@{\extracolsep{0.5cm}}cc}
    \toprule
      Model   &  PSNR$\uparrow$(dB)  &  SSIM$\uparrow$  \\
    \midrule
      \textit{Net1} & 26.14  & 0.801  \\
      \textit{Net2} & 26.76  & 0.826  \\
      \textit{Net3} & 27.20  & 0.837  \\
      \midrule
      Ours~($M=2$)  & 27.41  & 0.836  \\
      Ours~($M=9$)  & \textbf{28.56}  & \textbf{0.855}  \\
      Ours~($M=16$) & 27.98  & 0.850  \\
    \bottomrule
    \end{tabular}
    \caption{Ablation study of different warping methods. }
    \label{tab:ab_awm}
\end{table}

\begin{figure}[!htbp]\footnotesize
    \vspace{-2mm}
    \centering
    \begin{tabular}{@{\hspace{1mm}}c@{\hspace{0.5mm}}c@{\hspace{1mm}}}
        \includegraphics[width=0.48\linewidth]{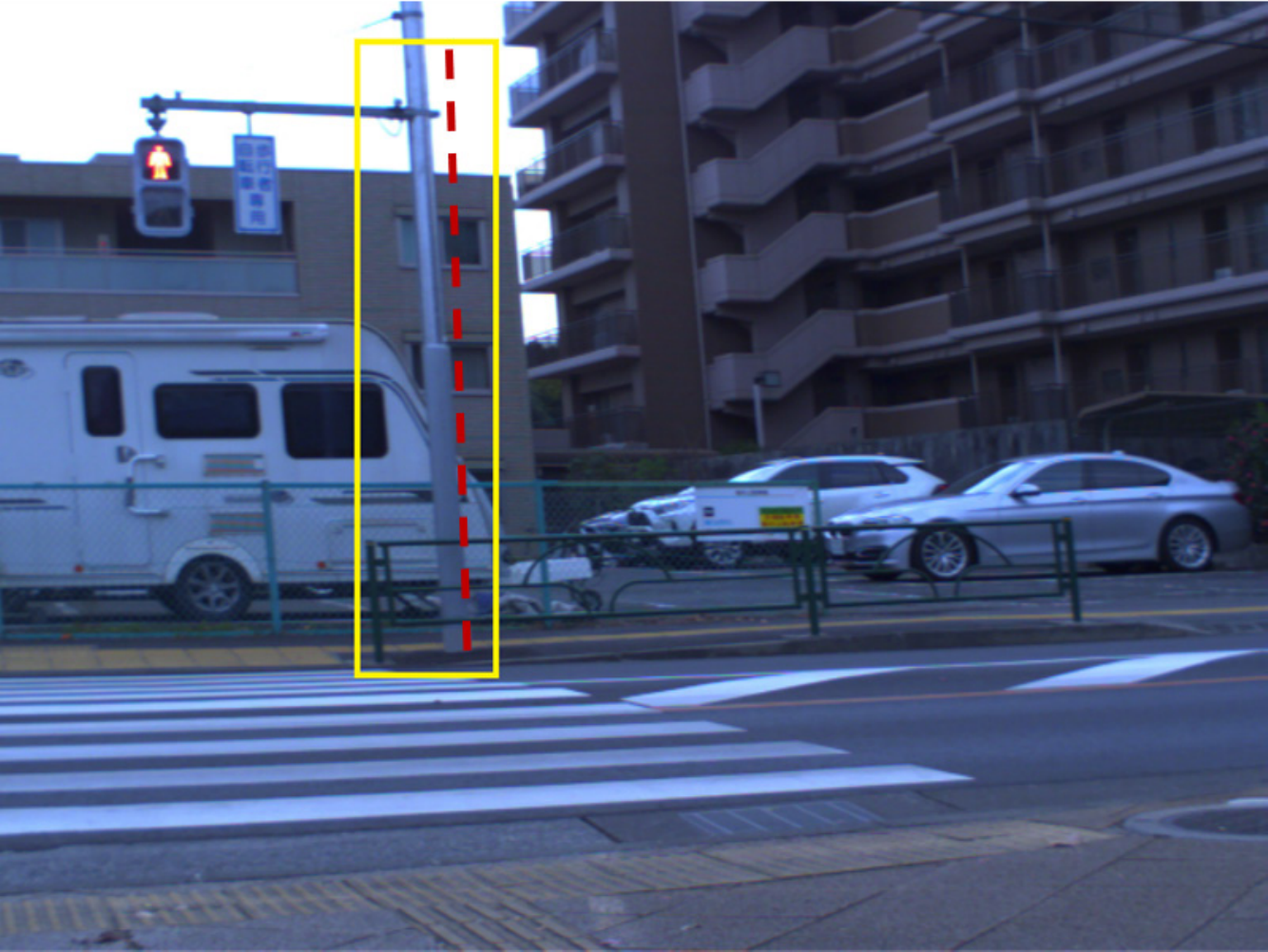} &
        \includegraphics[width=0.48\linewidth]{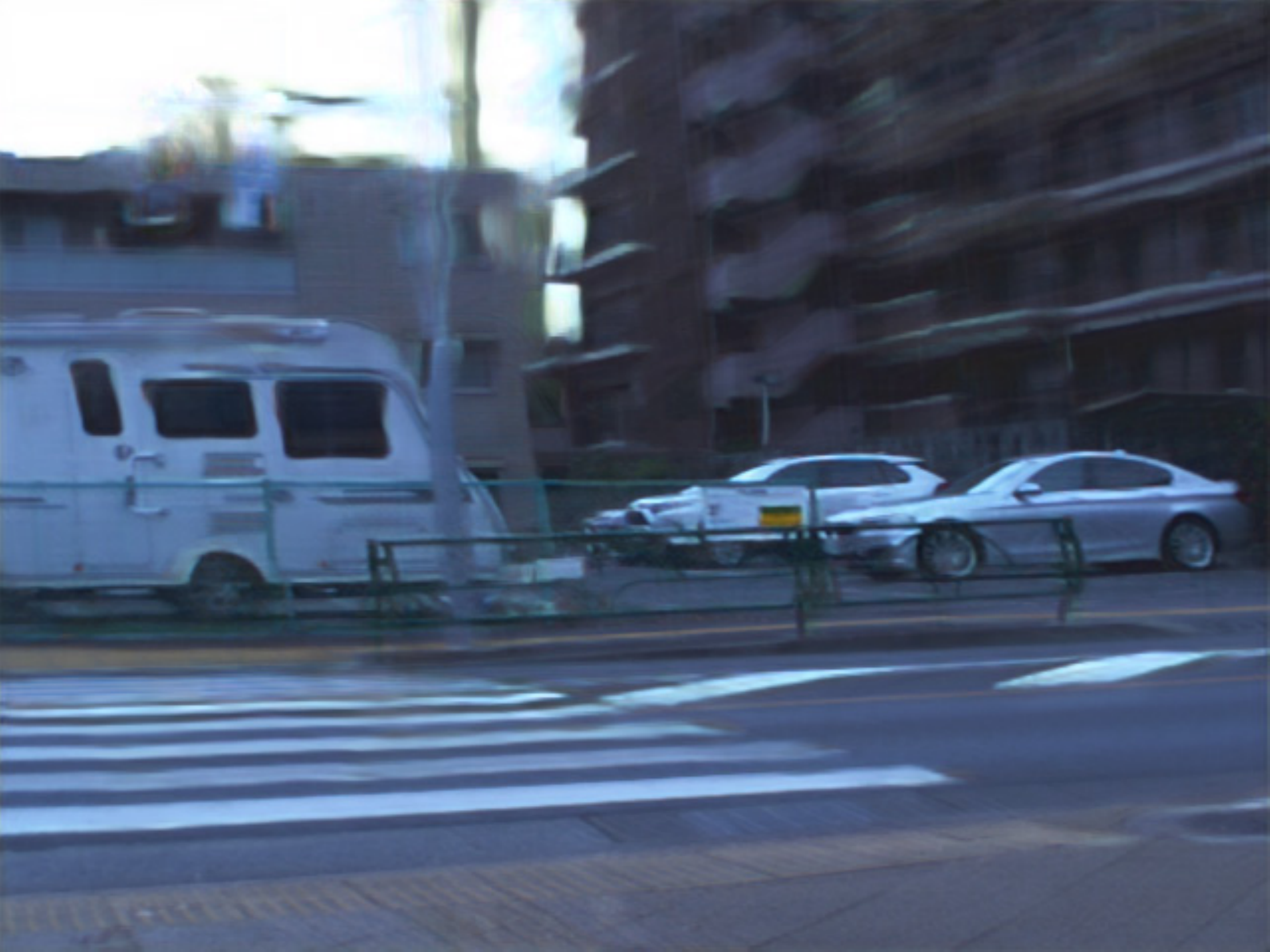}  \\
        (a) RS frame  & (b) Trained on Fastec-RS \\
        \includegraphics[width=0.48\linewidth]{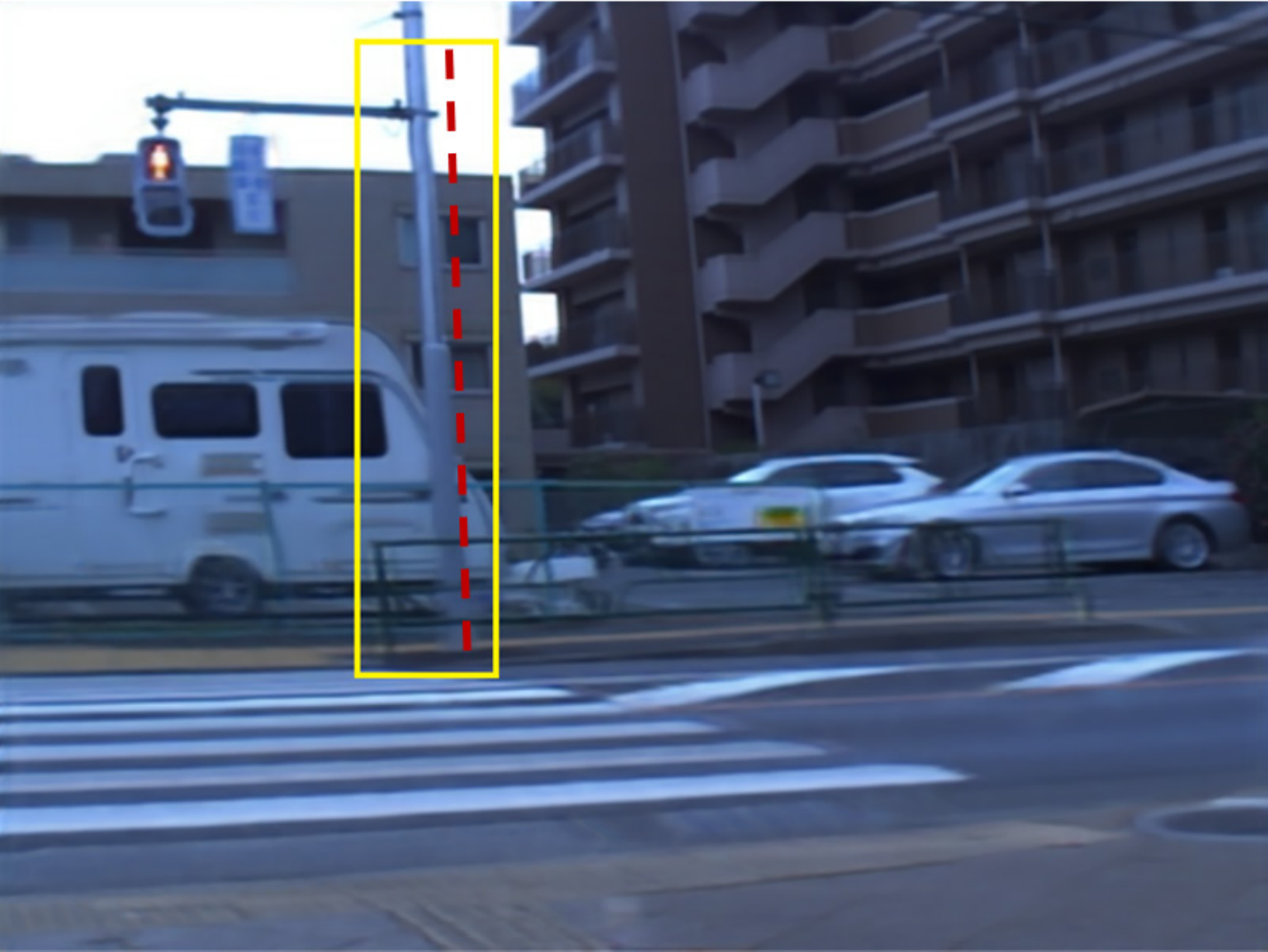} &
        \includegraphics[width=0.48\linewidth]{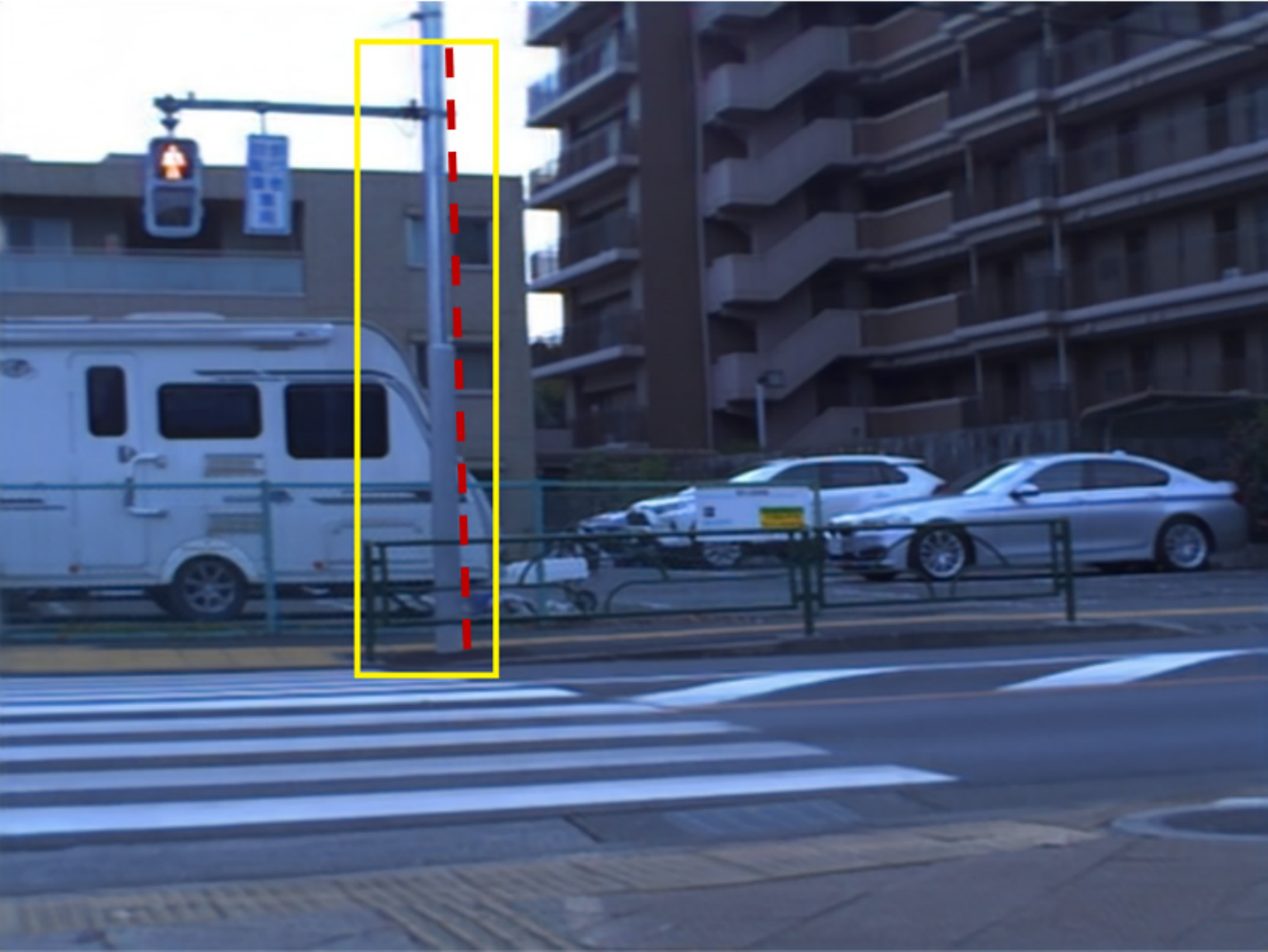}  \\
        (c) DSUN  &  (d) Ours \\
        \end{tabular}
        \caption{The corrected results of the proposed method on the frames captured by a third-party camera. \textbf{(a)} is the input RS frame. \textbf{(b)} is restored by our model trained on synthesized Fastec-RS. \textbf{(c)} and \textbf{(d)} are corrected by DSUN and our model trained on the proposed BS-RSC.} %
        \vspace{-5mm}
    \label{fig:results_cross_camera}
\end{figure}

\vspace{-2mm}
\noindent \textbf{Cross Camera Validation.}
To further validate the effectiveness of the proposed real-world RSC dataset BS-RSC, we test our model and DSUN model on the RS frames captured by third-party RS camera EO-1312C. The visual results are shown in \cref{fig:results_cross_camera}. The sub-figure (b) losses many details compared to the original RS frame, which shows that the model trained on the synthesized dataset Fastec-RS cannot remove the RS effects and even introduces more blurs and artifacts into the image. Sub-figure (c) and (d) demonstrate that the proposed BS-RSC can help deal with real-world RS distortions. However, the DSUN model cannot estimate the displacement field effectively and correct the RS frame well. Thanks to the design of adaptive warping, our model obtains visually friendly results.

\begin{figure}[!htbp]\footnotesize
    \centering
    \includegraphics[width=\linewidth]{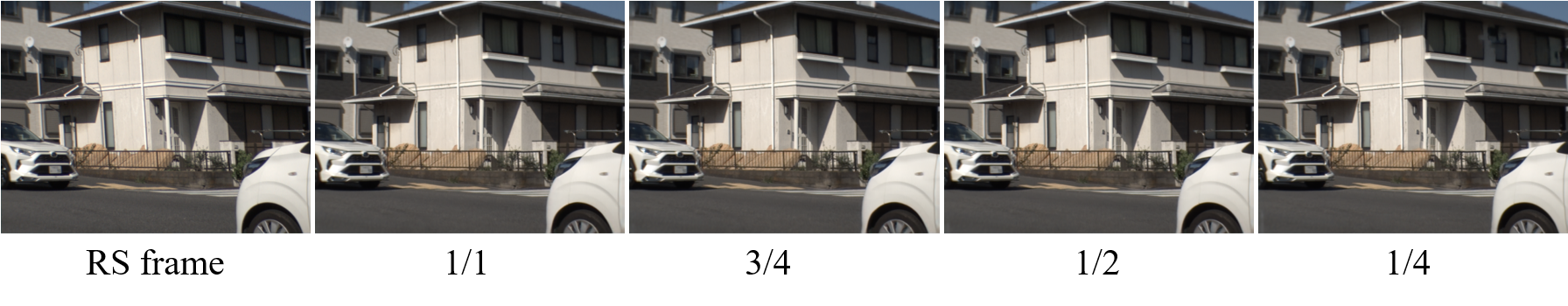}
    \vspace{-5mm}
    \caption{The corrected results of RS frames with different inter-frame time. } %
    \label{fig:inter_frame_time}
    \vspace{-3mm}
\end{figure}

\vspace{2mm}
\noindent\textbf{Inter-frame Time.}
To validate the generalisation capability of the proposed model, we test the trained model on the RS videos with different inter-frame time (by interpolating the RS frames), and the corrected results at different time stamps are shown in \cref{fig:inter_frame_time}. We see that our model is robust to different inter-frame time of the input RS frames during testing. However, some minor artifacts will occur (\eg, the corrected GS frame with 1/4 inter-frame time) when the testing inter-frame time largely deviates from the inter-frame time of the training dataset. Therefore, we cannot perfectly avoid overfitting the inter-frame time of training dataset.

\vspace{-4mm}
\section{Limitation and Conclusion}
\vspace{-1mm}
In this paper, we explore the real-world rolling shutter correction task. An effective adaptive warping model based on the attention mechanism is proposed, and a real-world RSC dataset is collected by a well-designed beam-splitter acquisition system. Experimental results demonstrate the effectiveness of both, showing highly comparative results against previous warping-based methods. However, real-time inference on low-power mobile devices is still challenging at now, and how to accelerate the model is our future work. 

\noindent\textbf{Acknowledgments.} This work was supported partially by the Major Research Plan of the National Natural Science Foundation of China (Grant No. 61991450), the Shenzhen Key Laboratory of Marine IntelliSense and Computation under Contract ZDSYS20200811142605016, and the JSPS KAKENHI with Grant Number 20H05951.

{\small
\bibliographystyle{ieee_fullname}
\bibliography{egbib}
}

\end{document}